\documentclass{article}
\usepackage{iclr2027_conference,times}
\usepackage{amsmath,amssymb,amsthm}
\usepackage{graphicx,adjustbox}
\usepackage{array,tabularx,booktabs,makecell}
\usepackage{enumitem,capt-of,titlesec}
\usepackage[most]{tcolorbox}
\usepackage{newunicodechar}
\newunicodechar{♫}{}
\usepackage{hyperref}
\hypersetup{pdfauthor={}}

\usepackage{amsmath,amsfonts,bm}

\def\eqref#1{equation~\ref{#1}}

\def\1{\bm{1}}

\DeclareMathAlphabet{\mathsfit}{\encodingdefault}{\sfdefault}{m}{sl}
\SetMathAlphabet{\mathsfit}{bold}{\encodingdefault}{\sfdefault}{bx}{n}

\renewcommand{\eqref}[1]{\textup{(\ref{#1})}}
\usepackage{placeins}

\theoremstyle{definition}

\newtheorem{definition}{Definition}
\newtheorem{theorem}{Theorem}
\newtheorem{assumption}{Assumption}

\newtheoremstyle{appendixplain}
  {5pt}
  {2pt}
  {\itshape}
  {0pt}
  {\bfseries}
  {.}
  {0.5em}
  {\thmname{#1}\thmnumber{ #2}\thmnote{ (#3)}}

\theoremstyle{appendixplain}

\newtheorem{applemma}{Lemma}
\newtheorem{appproposition}{Proposition}

\newtcolorbox{matchedexample}[1]{
  colback=black!2,
  colframe=black!35,
  colbacktitle=black!7,
  coltitle=black,
  title=\textbf{#1},
  fonttitle=\small\bfseries,
  fontupper=\small,
  boxrule=0.4pt,
  arc=0pt,
  left=5pt,
  right=5pt,
  top=4pt,
  bottom=4pt,
  before skip=4pt,
  after skip=4pt
}

\title{A Mechanistic Analysis of In-Context Learning and Out-of-Context Hallucination}
\author{Tianxiang Chen, Jingyuan Zhou, Guoyao Yang, Kaidi Yang\thanks{Corresponding author.} \\
Department of Civil and Environmental Engineering \\
National University of Singapore \\
\texttt{\{e1373675,jingyuanzhou,guoyao.yang\}@u.nus.edu}, \texttt{ykaidi@nus.edu.sg}}

\iclrfinalcopy

\begin{document}
\maketitle
\begin{abstract}
Large language models can infer and follow specific rules through in-context learning (ICL), yet may fail to remain faithful to provided evidence, giving rise to out-of-context (OOC) hallucination. Existing work has largely investigated the underlying mechanisms of ICL and OOC hallucination along separate lines. However, this separation obscures the mechanistic basis for these seemingly opposing outcomes. In this work, we study this contrast through the unified view of functional roles in dynamic context. The key observation is that role information can be internally represented without reliably constraining output. We derive a first-order decomposition to characterize the role effect, showing that its substantial influence depends on two necessary conditions, namely sufficient contribution magnitude and coherent signed aggregation. Causal analysis further shows that role span contributions affect key token choices involving response structure, content selection, and factual values. We further show that role-consistent choices depend on sufficient role span support at these key decisions, while failing choices can be reversed when this support overcomes the deficit. These findings support a common mechanistic view of ICL and OOC hallucination: each functional role defines the pattern of output decisions over which role span contributions must remain effective, and differences in this organization account for the seemingly opposing outcomes.
\end{abstract}

\section{Introduction}
A central strength of LLMs is the ability to condition generation on dynamic instructions. This capability is exemplified by in-context learning (ICL), where models can infer and follow specific rules supplied in the prompt~\citep{brown2020language,dong2024survey}. The same dependence on context, however, also exposes a major reliability challenge. Models may fail to remain faithful to the provided evidence and even produce unsupported claims, a failure mode we refer to as out-of-context (OOC) hallucination~\citep{huang2025hallucination,wu2024clasheval}. 
This contrast raises a natural question: why can the same model reliably follow a prompt-specified rule, yet fail to remain faithful to evidence that is likewise supplied in context?
Although existing work studies the mechanisms underlying ICL and OOC hallucination separately~\citep{hendel2023task,todd2024function,yu2024mechanistic}, the mechanistic origin of this contrast remains unclear. 

Fundamentally, dynamic context serves different functional roles in generation: rules specify how to respond, whereas evidence specifies what factual content the response should be grounded in. The same content can be interpreted differently according to its function in the current prompt. The same sentence may act as a rule, a distractor, or ordinary background across tasks~\citep{shi2023distracted}. While contemporary LLMs receive no explicit supervision for these roles, recent mechanistic studies of ICL suggest that task structure can nevertheless be inferred from context~\citep{cho2025revisiting}.

In this paper, we present a counter-intuitive finding: role information can be internally represented without reliably constraining the resulting response. We decompose this distinction into two questions: (1) how functional roles are internally represented, and (2) how role information affects LLM generation. Despite their different outcomes, ICL and OOC hallucination can be analyzed through a unified framework focusing on the role effect, while preserving the distinct functions of rules and evidence. We connect these questions by examining how information from role-bearing spans affects token predictions and choices during generation. Crucially, a role can affect the model's predictions without determining its final choice.

The effect of dynamic context is inherently non-uniform~\citep{an2024utilize}. The prefix can establish the direction of the continuation and persistently influence subsequent predictions. For instance, rules can shape how the response is organized, whereas evidence can determine which factual content is carried forward~\citep{xie2024adaptive,du2024context,wu2025retrieval}. We isolate the attention contributions from role-bearing spans, termed role span contributions, and examine how they affect next-token predictions. Our experiments show that successful role-conditioned generation depends on sufficient contribution support at the relevant token decisions. From this perspective, we show that ICL and OOC hallucination share the same contextual-influence mechanism, while it takes effect at different types of decisions in the two settings. Interested readers can refer to Appendix~\ref{app:related-work} for a thorough discussion of related work.

Our contributions are threefold. First, we establish a unified account of contextual influence in ICL and OOC hallucination by distinguishing internal role representation from its effect on generation. Second, we derive a first-order decomposition that identifies two necessary conditions for predictive influence: sufficient contribution magnitude and coherent signed aggregation. Third, we show that ICL and OOC hallucination share a role span contribution mechanism, while their contrasting outcomes depend on effective support at different generation decisions.

\section{Preliminaries}
\label{sec:preliminaries}

\textbf{Setup.} Let $C$ denote the context, $x$ the query, and $y=(y_1,\ldots,y_T)$ a response trajectory. In ICL, the context supplies a rule for answering the query, whereas in OOC hallucination, it supplies evidence against which the response is evaluated. We index two functional roles, rule and evidence, by $f\in\{r,e\}$, and use $C_f$ to denote the corresponding role span. For an input $P$, let $\mathcal S_f(P)$ denote its token positions. At decoding step $t$, the model conditions on the prefix $y_{<t}$ and defines a next-token distribution $p_\theta(\cdot \mid C,x,y_{<t})$. We refer to this next-token prediction as the decision at step $t$, and to $y_t$ as its choice. As the role span sits in the prompt, its effect on generation is realized through these decisions.

\begin{figure}[htbp]
\centering
\includegraphics[width=0.8\linewidth]{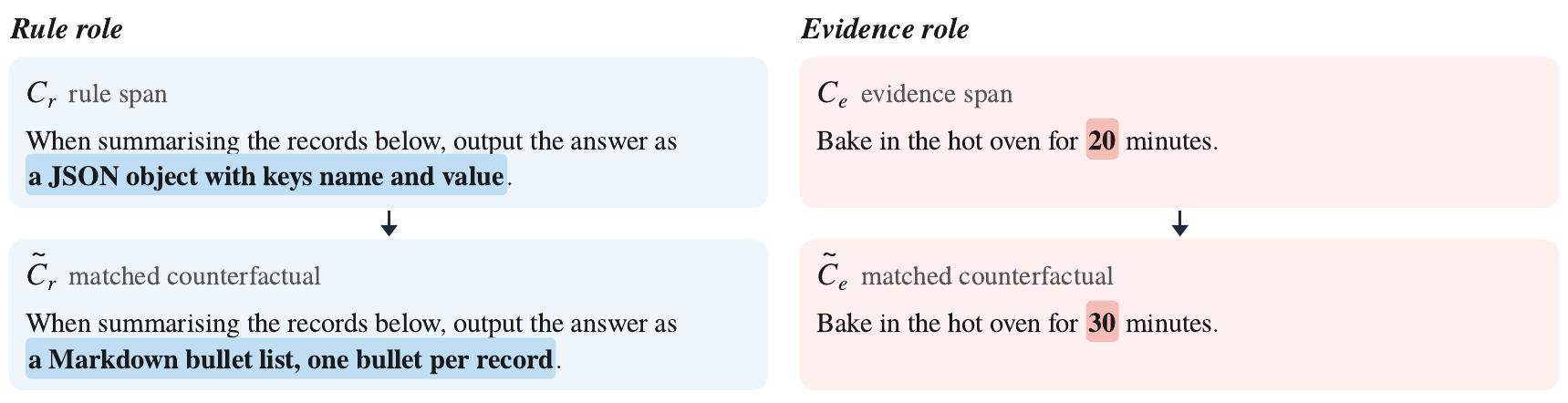}
\caption{\textbf{Examples of matched contexts}. Construction details are provided in Appendix~\ref{app:matched-role-construction}.}
\label{fig:matched-roles}
\end{figure}

\textbf{Matched contexts.} To isolate the effect of role \(f\), we construct a matched counterfactual context \(\widetilde C_f\) that changes only the content of the role span while holding the remaining prompt fixed, as illustrated in Figure~\ref{fig:matched-roles}. The role \(f\) itself is unchanged, and only what it specifies differs. We write \(P^f\) and \(\widetilde P^f\) for the inputs constructed from \(C_f \oplus x\) and \(\widetilde C_f \oplus x\), where \(\oplus\) denotes sequence concatenation. 

\textbf{Residual response.} Inspired by prior work on LLM internal mechanisms~\citep{todd2024function}, we investigate how a role span is represented in the model and how it affects the internal state. Let \(h_\ell(P,i)\in\mathbb R^d\) denote the residual stream representation at layer \(\ell\) and position \(i\) when the model processes \(P\). We define the residual response as \(\Delta h_\ell^f(i) = h_\ell(P^f,i)-h_\ell(\widetilde P^f,i)\), which points from the matched counterfactual state to the original role instantiation.

\section{Role Spans Induce Functional Residual Responses}
\label{sec:shared-representation}
ICL and OOC hallucination differ in their empirical outcomes, yet both rely on dynamic context within the same autoregressive decoding framework. Despite the shared structure, context serves distinct functional roles in the two settings. This raises a natural question: \textbf{\textit{how does the functional role organize the model’s internal state}}? We address this through matched role comparisons in the residual stream, linking the role distinction to its internal representational form. We first characterize the organization of the induced state changes and then examine their relation to the decoding process.

\subsection{Structured Residual Responses from Role Spans} 

The same sentence can serve different functional roles, such as an instruction, evidence, or distractor. To examine how this distinction appears in internal states, we keep the span content fixed while changing only its function. Specifically, we construct multiple functional framings for each span and train a linear readout to distinguish the corresponding roles, following \citet{belinkov2022probing}. We find that the readout generalizes consistently across prompt formulations, indicating a stable functional distinction in internal states.

\begin{figure}[htbp]
    \centering
    \setlength{\abovecaptionskip}{3pt}
    \setlength{\belowcaptionskip}{0pt}
    \includegraphics[width=0.75\linewidth]{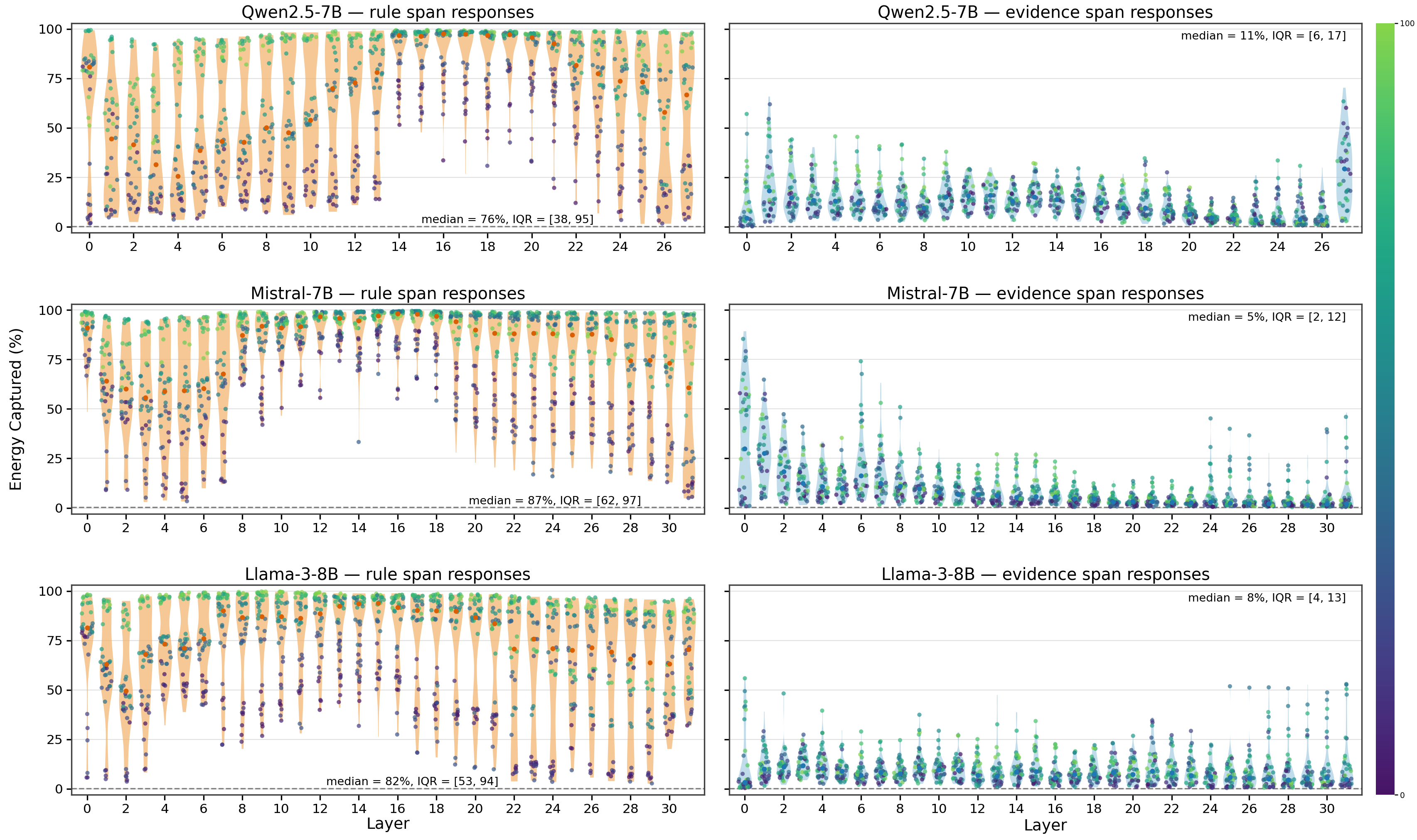}
    \caption{ \textbf{Structured residual responses}. Left panels show rule span responses, and right panels show evidence span responses. Each violin shows the fraction of $\Delta h$ energy captured by a 10-dimensional PCA subspace; points are colored by the percentile of $\|\Delta h\|$. The dashed gray line denotes the expected random subspace baseline \(k/H \approx 0.2\%\), with \(k=10\) PCA directions and residual dimension.}
    \label{fig:structured_residual_responses}
    \vspace{-4pt}
\end{figure}

We next examine how functional role structures internal states. To test this, for each role \(f\), we fit the top 10 PCA directions on one half of the samples and measure the fraction of $\Delta h_\ell^f$ energy captured on the other half. Ideally, a high captured fraction would indicate shared organization, whereas isotropic differences would remain near the random baseline. Figure~\ref{fig:structured_residual_responses} shows that the energy concentration remains well above the baseline. Rule responses exhibit strong concentration, whereas evidence responses are more distributed. These results show that the residual responses induced by functional role exhibit systematic low-dimensional organization.

\subsection{Residual Responses Toward Role-Consistent Generation}

The low-dimensional organization observed above shows that role spans can induce structured representational change during inference. Since the residual stream connects prompt processing to token prediction, we next examine how these changes relate to the output process. To make this relation concrete, we consider two empirical implications: (1) \textbf{The dominant directions of the residual subspace should align with the unembedding directions of outputs.} This follows from the fact that the LLM maps the residual representation to token logits through the final normalization and unembedding matrix $W_U$. Here, the structured $\Delta h$ should appear as a larger projection onto the output directions. (2) \textbf{The representation change induced by a specific role span should preserve relevant role information.} Since the residual stream connects intermediate representations with downstream computation, the role information represented by $\Delta h$ should be maintained in a form that later blocks can read, transform, and propagate.

\begin{figure}[htbp]
\centering

\begin{minipage}[t]{0.515\textwidth}
\vspace{0pt}
\centering

\includegraphics[
    width=0.98\linewidth,
    trim=0pt 0pt 0pt 0pt,
    clip
]{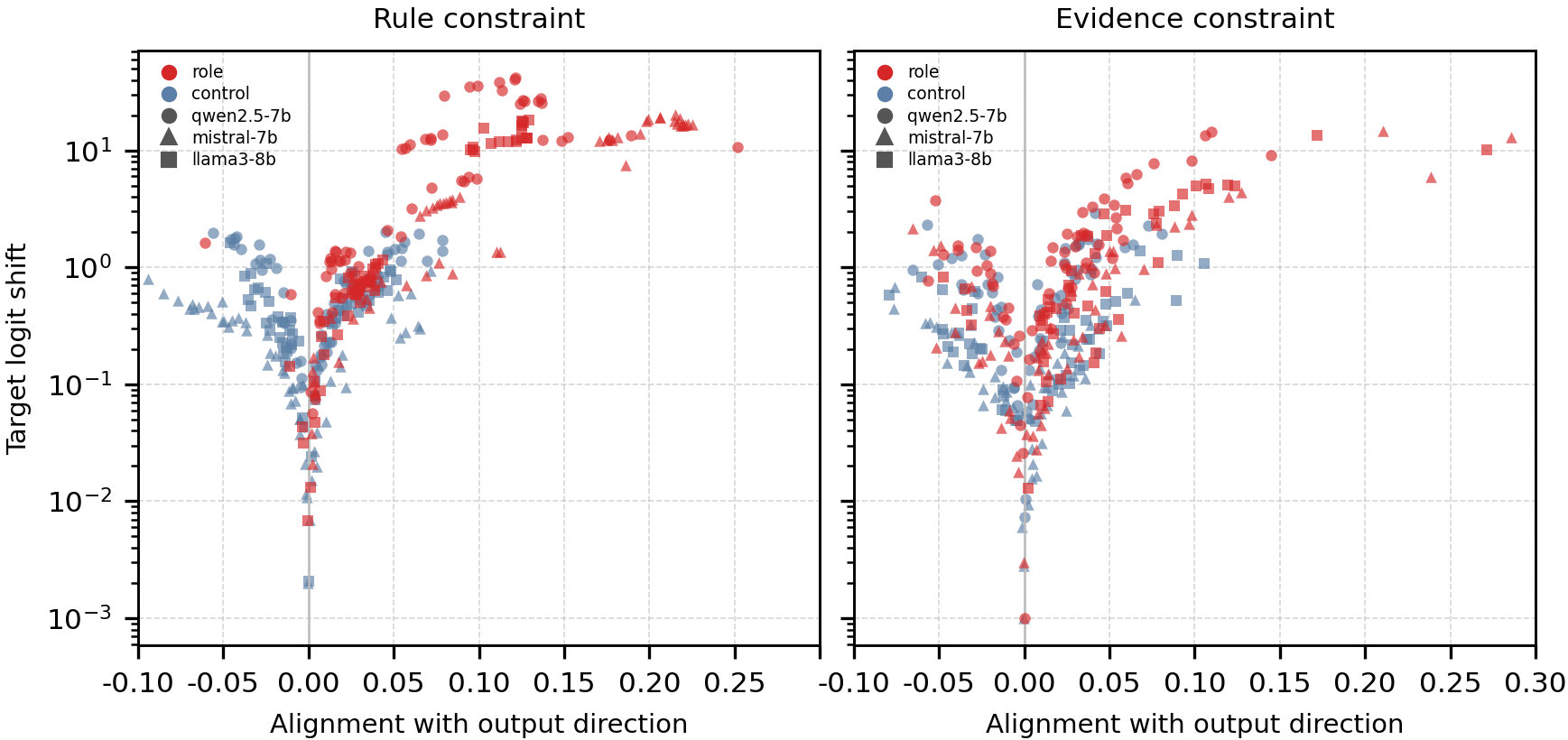}

\vspace{5pt}

\begin{minipage}{0.98\linewidth}
\centering
\setlength{\abovecaptionskip}{0pt}
\setlength{\belowcaptionskip}{0pt}
\captionof{figure}{
\textbf{Final layer decoding projection of $\Delta h$.}
Blue points denote the comparison group.}
\label{fig:p4.1}
\end{minipage}
\end{minipage}%
\hfill
\begin{minipage}[t]{0.465\textwidth}
\vspace{0pt}
\centering

\setlength{\abovecaptionskip}{0pt}
\setlength{\belowcaptionskip}{4pt}
\captionof{table}{\textbf{Behavior recovery rates}.}
\label{tab:p4.2}
\vspace{1pt}
\scriptsize
\setlength{\tabcolsep}{2.4pt}
\renewcommand{\arraystretch}{0.94}
\resizebox{\linewidth}{!}{%
\begin{tabular}{@{}lcccc@{}}
\toprule
Injected response & Qwen & Mistral & Llama & Avg. \\
\midrule
\multicolumn{5}{@{}l}{\textit{Rule}} \\
matched role response
& \textbf{0.94} & \textbf{1.00} & \textbf{1.00} & \textbf{0.98} \\
no role content
& 0.17 & 0.08 & 0.00 & 0.08 \\
mismatched role response
& 0.10 & 0.31 & 0.05 & 0.15 \\
\midrule
\multicolumn{5}{@{}l}{\textit{Evidence}} \\
matched role response
& \textbf{0.71} & \textbf{0.80} & \textbf{0.93} & \textbf{0.81} \\
no role content
& 0.44 & 0.20 & 0.37 & 0.34 \\
mismatched role response
& 0.35 & 0.25 & 0.04 & 0.21 \\
\bottomrule
\end{tabular}%
}
\end{minipage}
\vspace{-2pt}
\end{figure}

To examine the first empirical implication, we project $\Delta h$ through \(W_U\), measuring its projection and logit shift toward the output. We use a matched non-role perturbation as the comparison group. Figure~\ref{fig:p4.1} shows that, across all three models, role responses are more strongly aligned with the corresponding output directions, which results in larger logit shifts. Specifically, rule responses show clearer positive projection, while evidence responses separate more through the logit shift. By contrast, the comparison group remains weaker in both projection and logit shift. These results show that, during the decoding process, structured $\Delta h$ appears as a directional logit shift toward the output direction indicated by the role span.

To examine the second empirical implication, we evaluate behavior recovery in states induced by the matched counterfactual context; details are provided in Appendix~\ref{app:representation-recovery}. As shown in Table~\ref{tab:p4.2}, matched role responses restore the original behavior associated with the role, showing that the injected residual response remains usable by subsequent computation. Moreover, recovery remains low for the comparison groups since they lack the role conditioned state required for recovery. These results indicate that the induced representational change carries transferable role information that remains usable by downstream computation.

Our findings suggest that functional roles in ICL and OOC hallucination can be characterized in the representation space. Rule and evidence spans induce a common form of structured residual response that carries role-specific information forward for downstream computation.

\section{Role Span Contributions in Generation}
So far, we have characterized how role spans are internally represented, yet how they shape LLM generation remains unclear. We examine two questions: (i) whether role information can shift the model's preference toward specific responses, and (ii) how role spans affect token decisions during generation, leading to distinct generation outcomes. Specifically, we develop a theoretical framework that relates role-induced changes to internal LLM computations and derive a first-order decomposition to characterize the conditions under which role span contributions can influence LLM behavior.

\subsection{Response Trajectory Score Shift}
To characterize how rule or evidence spans influence generation, we examine their effects on next-token scores. Since autoregressive predictions depend on the preceding response history, changes in role information can lead to different response prefixes during generation. Such differences can confound the influence of role information on subsequent predictions. To isolate the influence of a role span at a given position, we evaluate the same token choice under a common response history. For each role \(f\in\{r,e\}\), we consider a response trajectory \(y=(y_1,\ldots,y_T)\), which can be either a constructed candidate or a response generated autoregressively. In both cases, we evaluate the score of \(y_t\) under contexts \(C_f\) and \(\widetilde C_f\), holding the query \(x\) and response prefix \(y_{<t}\) fixed. We defer further details to Appendix~\ref{app:matched-role-construction}. Specifically, we write the stepwise score as \(\psi_t(P)=\log p_\theta(y_t\mid P)\), and define the resulting score shift as \(\delta_t^f(y)=\psi_t(P_t^f(y))-\psi_t(\widetilde P_t^f(y))\). Based on the autoregressive factorization, the change in the log probability assigned to \(y\) is the sum of these stepwise differences. We formalize this quantity in the following definition.

\begin{definition}[Response trajectory score shift]
For a response sequence \(y=(y_1,\ldots,y_T)\), define the stepwise score shift as
\(\delta_t^f(y)=\psi_t(P_t^f(y))-\psi_t(\widetilde P_t^f(y))\).
The response trajectory score shift is
\(\Phi_f(y)=\sum_{t=1}^{T}\delta_t^f(y)\),
which measures the change in the log probability assigned to the same response under \(C_f\) and \(\widetilde C_f\).
Equivalently, using the autoregressive product form,
\begin{equation}
\begin{aligned}
\Phi_f(y)
&=
\log p_\theta(y\mid C_f \oplus x)
-
\log p_\theta(y\mid \widetilde C_f \oplus x)
\\
&=
\sum_{t=1}^{T}
\left[
\log p_\theta(y_t\mid C_f \oplus x \oplus y_{<t})
-
\log p_\theta(y_t\mid \widetilde C_f \oplus x \oplus y_{<t})
\right].
\end{aligned}
\label{eq:response-trajectory-score-shift-factorized}
\end{equation}
\end{definition} Here, \(\Phi_f(y)>0\) indicates that \(C_f\) yields a higher score than \(\widetilde{C}_f\) does. In what follows, we relate the stepwise score shifts $\delta_t^f(y)$ to forward pass quantities associated with the role span.

\subsection{Role Span Contributions in the Forward Pass}
\label{sec:attn-output-source-decomposition}

We mainly consider decoder-only LLMs, where successive attention and MLP modules update the residual stream before the final unembedding. The role-induced responses observed in Sec.~\ref{sec:shared-representation} are formed and propagated through these updates. To relate this process to the role span, we write the residual update explicitly. For a processed input \(P\), the residual state at layer \(\ell\) and position \(i\) is updated as \(h_\ell(P,i)=h_{\ell-1}(P,i)+\operatorname{Attn}^{\mathrm{out}}_\ell(P,i)+\operatorname{MLP}^{\mathrm{out}}_\ell(P,i)\). This reveals that information carried by different roles is accumulated across depth. A difference induced by role \(f\) can therefore reflect both information inherited from previous layers and the current updates. Recall that the role span \(\mathcal S_f(P)\) is defined by source token positions. Among the two update terms above, the attention output can be explicitly written as a sum over source token positions: 
\begin{equation}
\operatorname{Attn}^{\mathrm{out}}_\ell(P,i)
=
\sum_a\sum_s
\alpha^{(a)}_{i\leftarrow s}(P)
V_\ell^{(a)}(P,s)W_O^{(a)}
\label{eq:attn-output-source-decomposition}
\end{equation}  where \(a\) indexes attention heads and \(s\) indexes source token positions in \(P\) that are visible to the target position \(i\) under the causal mask. Eq.~\eqref{eq:attn-output-source-decomposition} decomposes the attention update at \(i\) into contributions aggregated from source positions. The key observation is that the role span \(\mathcal S_f(P)\) is also a set of input token positions. We can therefore isolate the contribution of the role span by retaining the terms with \(s\in\mathcal S_f(P)\), giving \begin{equation}
B_{\ell,i}^f(P)
=
\sum_a
\sum_{s\in\mathcal S_f(P)}
\alpha^{(a)}_{i\leftarrow s}(P)
V_\ell^{(a)}(P,s)W_O^{(a)}.
\label{eq:role span-attention-contribution}
\end{equation} 
We term \(B_{\ell,i}^f(P)\) the role span contribution to the attention update at layer $\ell$ and position \(i\). In decoder-only Transformers, attention directly transfers differences across token positions, whereas normalization and MLP blocks act only within individual positions. Appendix~\ref{app:attention-contribution} formalizes this distinction.

\subsection{First-Order Decomposition of Score Shifts}
\label{sec:role span-contributions}
The response trajectory score shift \(\Phi_f(y)\) accumulates the stepwise shifts \(\delta_t^f(y)\) across response positions. We therefore analyze each stepwise shift through selected role span contribution coordinates and quantify the first-order score change associated with their matched differences. To make this operational, we approximate the stepwise score by its first-order change with respect to the selected \(B_{\ell,i}^f\) coordinates. Specifically, for each decoding step \(t\), let \(\mathcal B_{f,t}\subseteq\{(\ell,i)\}\) denote the role span contribution indices considered in the expansion. For \((\ell,i)\in\mathcal B_{f,t}\), define \(\Delta B_{\ell,i,t}^f(y)
= B_{\ell,i}^f(P_t^f(y)) - B_{\ell,i}^f(\widetilde P_t^f(y))\). Collecting these differences gives
\(
\Delta B_{\mathcal B,t}^f(y) =
\left(
\Delta B_{\ell,i,t}^f(y)
\right)_{(\ell,i)\in\mathcal B_{f,t}} \). The vector $\Delta B^f_{\mathcal{B},t}(y)$, therefore, measures how the matched role comparison changes the selected role span contributions over $\mathcal{B}_{f,t}$.
\begin{assumption}[Local score regularity]
\label{ass:local-score-regularity}
\textit{For a fixed response position \(t\), let \(\widetilde b\) and \(b^f\) denote the selected role span contribution vectors induced by \(\widetilde P_t^f(y)\) and \(P_t^f(y)\), respectively. Define \(\Psi_t(b)\) as the score obtained by replacing, in layer order, the selected contributions under \(\widetilde P_t^f(y)\) with the corresponding entries of \(b\). By construction, \(\Psi_t(\widetilde b)=\psi_t(\widetilde P_t^f(y))\); let \(E_t^\perp=\psi_t(P_t^f(y))-\Psi_t(b^f)\). We assume that \(\Psi_t\) is differentiable and has an \(L_t\)-Lipschitz gradient on the line segment from \(\widetilde b\) to \(b^f\), and write \(\nabla_{B_{\ell,i}^f}\psi_t:=\nabla_{b_{\ell,i}}\Psi_t(\widetilde b)\).}
\end{assumption} Under Assumption~\ref{ass:local-score-regularity}, a first-order expansion of \(\Psi_t\) from \(\widetilde b\) to \(b^f\) yields the following decomposition.

\begin{theorem}[First-order role span score decomposition]
\label{theorem:First-order role span score decomposition}
\textit{Suppose Assumption~\ref{ass:local-score-regularity} holds. Then the score shift satisfies}
\begin{equation}
\delta_t^f(y)
=
\sum_{(\ell,i)\in\mathcal B_{f,t}}
\left\langle
\nabla_{B_{\ell,i}^f}\psi_t,
\Delta B_{\ell,i,t}^f(y)
\right\rangle
+
R_t,
\label{eq:first-order-role span-score-shift}
\end{equation}
\textit{where the gradients are evaluated at \(\widetilde b\), and the remainder satisfies}
\begin{equation}
|R_t|
\le
\frac{L_t}{2}
\left\|
\Delta B_{\mathcal B,t}^f(y)
\right\|^2
+
|E_t^\perp|.
\label{eq:first-order-role span-score-shift-remainder}
\end{equation}
\end{theorem} The detailed derivations are provided in Appendix~\ref{app:local-score-decomposition}. Eq.~\eqref{eq:first-order-role span-score-shift} expresses the stepwise score shift as selected first-order terms plus a remainder. We use the \(B_{\ell,i}^f\) coordinates to isolate the direct source-indexed role span attention pathway, while effects outside these coordinates remain in \(R_t\). We define the selected first-order score contribution associated with \((\ell,i)\) as \begin{equation}
C_{\ell,i,t}^{B,f}
=
\left\langle
\nabla_{B_{\ell,i}^f}\psi_t,
\Delta B_{\ell,i,t}^f(y)
\right\rangle .
\label{eq:role span-directional-contribution}
\end{equation} Note that \(C_{\ell,i,t}^{B,f} \) is an inner product between the local score gradient and the matched contribution difference, so it gives a directional contribution to the score shift. Hence, Eq.~\eqref{eq:role span-directional-contribution} makes two factors explicit: (1) the matched difference \(\Delta B_{\ell,i,t}^f(y)\) captures the role-dependent change in the forward pass quantity, and (2) the gradient \(\nabla_{B_{\ell,i}^f}\psi_t\) captures the local sensitivity of the stepwise score to this quantity. Equivalently, writing the inner product in norm-angle form gives \begin{equation}
C_{\ell,i,t}^{B,f}
=
\|\nabla_{B_{\ell,i}^f}\psi_t\|\,
\|\Delta B_{\ell,i,t}^f(y)\|\,
\cos\angle
\left(
\nabla_{B_{\ell,i}^f}\psi_t,
\Delta B_{\ell,i,t}^f(y)
\right).
\label{eq:role span-two-factor-contribution}
\end{equation} To this end, we identify three distinct factors: the matched difference \(\Delta B_{\ell,i,t}^f(y)\), the local score sensitivity, and their directional alignment. Summing Eq.~\eqref{eq:first-order-role span-score-shift} over response positions and using Eq.~\eqref{eq:role span-directional-contribution} gives \begin{equation} 
\Phi_f(y)
=
\sum_{t=1}^{T}
\sum_{(\ell,i)\in\mathcal B_{f,t}}
C_{\ell,i,t}^{B,f}+R,
\quad R=\sum_{t=1}^{T}R_t.
\label{eq:trajectory-role span-score-contribution}
\end{equation} 

Together, Eqs.~\eqref{eq:role span-two-factor-contribution} and~\eqref{eq:trajectory-role span-score-contribution} show how local role span contributions organize into a trajectory-level score shift. Locally, the contribution magnitude depends on both the matched change in \(B_{\ell,i}^f\) and the sensitivity of the stepwise score, while the directional alignment determines whether the term supports or opposes the score shift. Across the generation, the signed terms must aggregate coherently rather than cancel. The first-order decomposition therefore identifies two conditions for a trajectory-level effect, namely \textbf{sufficient magnitude} and \textbf{coherent signed aggregation.}

\section{Experiments}
\label{sec:experiments}

In this section, we investigate the mechanism underlying the common dependence on context and the distinct generation outcomes of ICL and OOC hallucination. We use rule-following tasks derived from CL-bench \citep{dou2026clbench} and evidence-grounding examples from RAGTruth \citep{niu-etal-2024-ragtruth}, with settings provided in Appendix~\ref{app:experimental-details}. To explain these distinct outcomes, we address two related questions: how role information affects model predictions and how its contributions influence token selection during generation. Our experiments examine three claims: (i) that role span contributions produce role-consistent preference shifts through organized layerwise effects, (ii) that rule and evidence realize this influence through different types of key token decisions, and (iii) that successful choices depend on sufficient contribution support, while failed choices can be reversed when this support overcomes the decision deficit. Appendices~\ref{app:demonstration-icl} and~\ref{app:broader-evidence} extend the analysis to broader ICL and evidence-grounding settings.

\subsection{Role Span Contribution Mechanism }

\begin{figure}[htbp]
  \centering
  \includegraphics[
    width=0.9\columnwidth,keepaspectratio
  ]{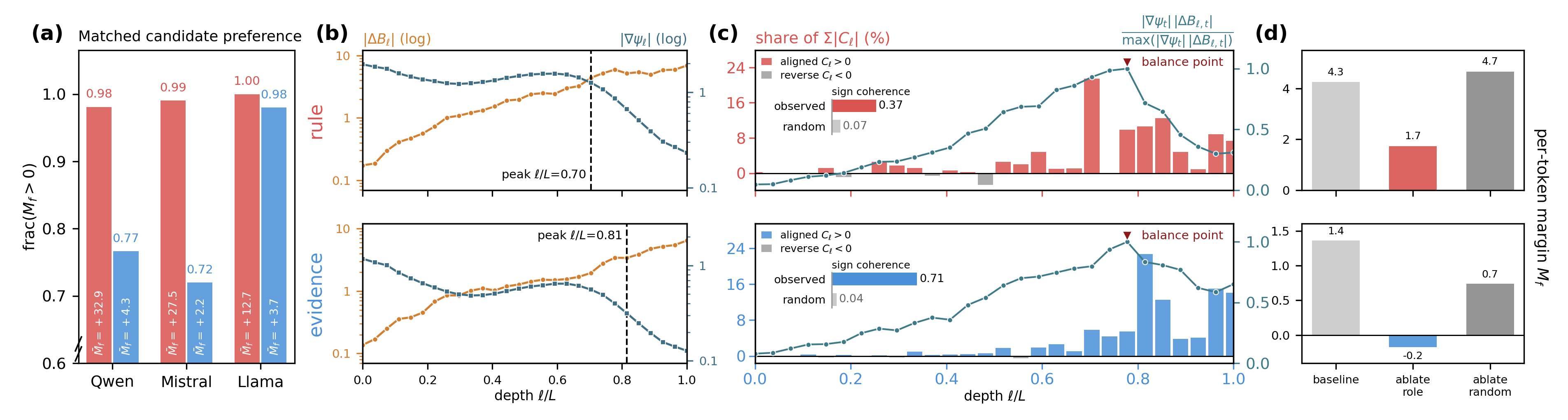}
  \caption{\textbf{Response preference and contribution organization}.
  \textbf{(a)} Fraction of positive candidate margins.
  For Qwen2.5-7B, \textbf{(b)} layerwise profiles of
  \( |\Delta B_{\ell,t}| \) and \( |\nabla \psi_t| \),
  \textbf{(c)} signed layerwise contributions and sign coherence,
  and \textbf{(d)} preference margin \(M_f\) under baseline, role span pathway ablation, and random ablation. More results are in Appendix~\ref{app:contribution-organization}.}
  \label{fig:qwen_composed}
\end{figure}

In this subsection, we empirically test how role span contributions translate into predictive influence as the first-order decomposition suggests. If a role span can guide predictions toward the specified content, it must increase the relative support for that content over alternatives. To examine this, we compare two candidate responses \(y^+\) and \(y^-\), where \(y^+\) follows the rule or is supported by the evidence, whereas \(y^-\) does not. Since we aim to measure how role information shifts predictive support between these alternatives, we evaluate their conditional probabilities under matched contexts, independently of the model's generated response. We define \(M_f=\Phi_f(y^+)-\Phi_f(y^-)\) to quantify the preference change. Here, $M_f>0$ indicates that role information increases the relative support for $y^+$ over $y^-$. Construction details are provided in Appendix~\ref{app:contribution-preference}. As shown in Figure~\ref{fig:qwen_composed}(a), we empirically observe that \(\mathrm{frac}(M_f>0)\) is high for both roles, indicating that the role span contribution consistently favors role-compatible responses over alternatives. More specifically, this preference is expressed as a tendency toward rule-following responses in ICL and evidence-grounded responses in the OOC setting.

Since the contribution propagates through the forward pass, its final effect accumulates across layers. The decomposition in Theorem~\ref{theorem:First-order role span score decomposition} suggests two conditions for a substantial first-order contribution: (1) the contribution magnitude must be sufficiently large, and (2) the signed layerwise terms must aggregate coherently. Based on Eq.~\eqref{eq:role span-two-factor-contribution}, we use $\|\nabla \psi_t\| \cdot \|\Delta B_{\ell,t}\|$ to measure contribution magnitude and the cosine term to determine the signed direction. Note that the aggregated first-order terms can track the exact score shift (Appendix~\ref{app:first_order_corr}).
                                                   
We first examine the layerwise contribution magnitude. As illustrated in Figure~\ref{fig:qwen_composed}(c), both rule and evidence exhibit a localized middle-to-late concentration with a clear interior peak, indicating that substantial contribution is confined to a limited depth range. To understand the mechanism behind this peak, we further decompose the contribution magnitude into its two factors. As shown in Figure~\ref{fig:qwen_composed}(b), $|\Delta B_{\ell,t}|$ and $|\nabla\psi_t|$ exhibit opposing trends: $|\Delta B_{\ell,t}|$ increases with depth, whereas $|\nabla\psi_t|$ decays. Given these opposing trends, their product is expected to peak at an intermediate depth where the increase in $|\Delta B_{\ell,t}|$ no longer outweighs the decrease in $|\nabla\psi_t|$. Consistent with this prediction, the resulting contribution exhibits the localized interior peak shown in Figure~\ref{fig:qwen_composed}(c). In early layers, $\|\Delta B_{\ell,t}\|$ remains close to zero, indicating that the role span has induced a limited change. The contribution therefore remains small despite high downstream sensitivity. Thus, a high contribution appears only when both factors remain sufficiently large. 

Notably, magnitude alone is insufficient if the signed terms cancel after aggregation. We further compare targeted contribution ablation with a random control, which controls for generic intervention effects. As shown in Figure~\ref{fig:qwen_composed}(d), targeted ablation produces a larger reduction in $M_f$ for both roles, showing that disrupting the contribution organization markedly weakens the resulting preference shift. Taken together, these results show that \textbf{role span contribution emerges when both magnitude factors remain sufficiently large}, while disrupting their signed organization weakens the resulting preference shift.

\subsection{Role Span Contribution at Key Token Decisions}
\label{sec:Role Span Influence} 

\textbf{Contribution localization.} The above analysis suggests that role spans can shift predictions toward role-consistent responses through layerwise contributions. During autoregressive generation, this influence must be realized through token choices that affect whether the role's requirements are satisfied. We therefore expect role span contributions to be particularly consequential at some key decisions. To examine how these decisions contribute to the distinct outcomes of ICL and OOC hallucination, we first measure the token-level first-order contribution \(\hat{\delta}_t\) at the corresponding output decisions. As shown in Table~\ref{tab:token_level_role_effect}(a), these decisions consistently occur at positions with high role span influence. We then remove the role span pathway during free generation and record the first change in token selection. These first divergences likewise occur at positions with high role span influence. Thus, these findings suggest that \textbf{role spans shape generation through key token choices.}

\textbf{Role Effects.} We next examine the effects of role span contributions at these key token decisions for the free generation. Specifically, we remove the role span pathway during decoding and assess whether the resulting response satisfies each role's requirements. As shown in Table~\ref{tab:token_level_role_effect}(b), removing the rule pathway disrupts the specified format and reduces retention of the selected entity. For evidence, pathway removal changes the facts stated in originally grounded responses. These results show that role span contributions causally support the key decisions that realize role-specific requirements. One might wonder whether the same pathway also affects factual decisions in real hallucinations. We test this using annotated LLM-generated hallucinations from RAGTruth~\citep{niu-etal-2024-ragtruth}, resuming generation from the prefixes preceding the hallucinated facts under the original and pathway-removed conditions. As shown in Table~\ref{tab:token_level_role_effect}(b), the rate of selecting evidence-supported facts drops from 61.5--62.3\% to 7.7--13.5\% across models. This finding extends the role effects observed above to factual decisions in real hallucinated generation.

Together, these results reveal distinct decision-level effects of the two roles. Rule span contributions are typically realized through structural and selection decisions, whereas evidence span contributions affect factual content decisions, including continuations from genuine hallucination examples. \textbf{By shaping different types of key token decisions, the two roles give rise to distinct generation outcomes in ICL and OOC hallucination.}

\begin{table}[htbp]
    \centering
    \setlength{\abovecaptionskip}{2pt}
    \setlength{\belowcaptionskip}{3pt}
    \caption{\textbf{Role span effects on output decisions.} \textbf{(a)} Role span influence at the first divergence and the output decision. \textbf{(b)} Retention rate of the content specified by $C_f$ under the original ($C_f$) and pathway-removed (Rem.) conditions. Additional results are provided in Appendix~\ref{app:role span-influence}.}
    \label{tab:token_level_role_effect}
    \footnotesize
    \setlength{\tabcolsep}{2.6pt}
    \begin{tabular}{@{}l cc cccc @{\hspace{10pt}} cccc cccc@{}}
        \toprule
        & \multicolumn{6}{c}{\textbf{(a) First-divergence localization}} & \multicolumn{8}{c}{\textbf{(b) Retention of role-specified output content (\%)}} \\
        \cmidrule(lr){2-7}\cmidrule(l){8-15}
        & \multicolumn{2}{c}{\textit{Rule span}} & \multicolumn{4}{c}{\textit{Evidence span}}
        & \multicolumn{4}{c}{\textit{Rule span}} & \multicolumn{4}{c}{\textit{Evidence span}} \\
        \cmidrule(lr){2-3}\cmidrule(lr){4-7}\cmidrule(lr){8-11}\cmidrule(l){12-15}
        & & & \multicolumn{2}{c}{Grounded} & \multicolumn{2}{c}{Hallucinated}
        & \multicolumn{2}{c}{Format} & \multicolumn{2}{c}{Entity}
        & \multicolumn{2}{c}{Grounded} & \multicolumn{2}{c}{Hallucinated} \\
        \cmidrule(lr){4-5}\cmidrule(lr){6-7}\cmidrule(lr){8-9}\cmidrule(lr){10-11}\cmidrule(lr){12-13}\cmidrule(l){14-15}
        Model & Attr. & Dec. & Attr. & Dec. & Attr. & Dec. & $C_f$ & Rem. & $C_f$ & Rem. & $C_f$ & Rem. & $C_f$ & Rem. \\
        \midrule
        Qwen2.5-7B   & 87.5 & 85.0 & 75.0 & 92.5 & 88.0 & 88.2 & 100.0 & 4.8 & 57.1 & 18.1 & 100.0 & 10.2 & 62.3 & 11.3 \\
        Mistral-7B   & 82.5 & 85.7 & 87.5 & 77.5 & 89.8 & 90.1 & 100.0 & 2.4 & 58.1 & 33.3 & 100.0 & 16.7 & 61.5 & 7.7 \\
        Llama-3.1-8B & 80.0 & 90.0 & 85.0 & 95.0 & 92.4 & 91.9 & 100.0 & 4.8 & 53.3 & 18.1 & 100.0 & 10.0 & 61.5 & 13.5 \\
        \bottomrule
    \end{tabular}
    \par\vspace{2pt}
    \begin{minipage}{0.95\columnwidth}
        \scriptsize
        \raggedright
        \textit{\textbf{Note}.} \textit{Attr.} and \textit{Dec.} report role span influence at the first divergence and output decision positions, respectively; higher values indicate stronger influence. Pathway removal changes token selection in 92--100\% of generations. Hallucinated prefixes are taken from RAGTruth. Further details are in Appendix~D.6.
    \end{minipage}
    \vspace{-4pt}
\end{table}

\subsection{Role Span Contributions to LLM Decision-Making}
\label{sec:Role Span Contributions to LLM Decision}

The preceding analyses establish how role span contributions are organized across layers and causally influence key token choices during free generation. Yet, even when role span contributions influence a decision, they may not be sufficient to make the model select the role-consistent token. We therefore examine when these contributions can sustain a successful choice or change an initially incorrect one. To this end, we first attenuate the signed contribution at initially successful decisions (Section~\ref{sec:experiments-3-1}) and then restore the contribution at naturally failing ones (Section~\ref{sec:experiments-3-2}).

\subsubsection{Dependence of Successful Decisions}
\label{sec:experiments-3-1}

We first examine successful decisions in the rule-following and evidence-grounding tasks derived above. In both settings, we analyze the model's autoregressive generations and identify the key decisions at which the corresponding role requirements are satisfied. To examine whether role span contributions sustain these choices, we measure the advantage of the role-consistent token over its strongest competitor using the target-token margin, with zero marking the decision boundary. We then attenuate the role span contribution by a factor of \(1-\alpha\) and track changes in this margin and the resulting token choice. 

As shown in Table~\ref{tab:decision_attenuation}(a), increasing $\alpha$ generally narrows the margin under both roles.  For rule decisions, the median margin shifts from positive to negative across all models, leaving the rule-specified token selected in only a minority of these key choices. Meanwhile, attenuation substantially erodes the advantage of the evidence-supported token, bringing the corresponding factual choice closer to the decision boundary. In contrast,  the non-role comparison shows smaller margin changes, showing that the pronounced changes are driven by role span contributions. Under full attenuation, these margin reductions are accompanied by frequent token flips and performance declines (Table~\ref{tab:decision_attenuation}(b).) 

To rule out dependence on predefined factual decisions, we further apply the same sustained attenuation to facts correctly generated by the models themselves. As shown in the response-derived rows of Table~\ref{tab:decision_attenuation}, targeted attenuation substantially reduces the decision margin and frequently changes the selected token and generated factual value, whereas attenuation of non-evidence spans largely preserves the original factual content. These results show that the presence of role information does not guarantee effective support during generation. \textbf{The contribution must remain sufficiently large when the model makes key choices.} For rules, this support must be sustained across repeated structural and selection decisions, whereas for evidence, it is required when the factual value is determined.

\begin{table}[htbp]
    \centering
    \setlength{\abovecaptionskip}{2pt}
    \setlength{\belowcaptionskip}{3pt}
    \caption{\textbf{Sustained attenuation of role span contributions}. The contribution is scaled by $(1{-}\alpha)$, with $\alpha{=}1$ denoting complete attenuation. \textbf{(a)} Median target token margin under targeted attenuation and non-role comparison. \textbf{(b)} Token flips and task performance under full attenuation.}
    \label{tab:decision_attenuation}
    \footnotesize
    \setlength{\tabcolsep}{2.4pt}
    \begin{tabular}{@{}l ccccc cc @{\hspace{8pt}} cc cc@{}}
        \toprule
        & \multicolumn{7}{c}{\textbf{(a) Target token margin}} & \multicolumn{4}{c}{\textbf{(b) Token and task outcomes at $\alpha{=}1$}} \\
        \cmidrule(lr){2-8}\cmidrule(l){9-12}
& \multicolumn{5}{c}{\makecell[b]{Targeted\\attenuation}} & \multicolumn{2}{c}{\makecell[b]{Non-role\\comparison}}
        & \multicolumn{2}{c}{\makecell[b]{Targeted\\attenuation}} & \multicolumn{2}{c}{\makecell[b]{Non-role\\comparison}} \\
        \cmidrule(lr){2-6}\cmidrule(lr){7-8}\cmidrule(lr){9-10}\cmidrule(l){11-12}
        Model & $\alpha{=}0$ & $\alpha{=}0.25$ & $\alpha{=}0.5$ & $\alpha{=}0.75$ & $\alpha{=}1$ & $\alpha{=}0$ & $\alpha{=}1$ & Flip (\%) & Task (\%) & Flip (\%) & Task (\%) \\
        \midrule
        \multicolumn{12}{@{}l}{\emph{Rule span}} \\
        Qwen2.5-7B   & 4.62 & 5.00 & -0.38 & -12.89 & -16.25 & 4.62 & 5.00 & 84.0 & 22.3 & 17.0 & 100.0 \\
        Mistral-7B   & 6.69 & 6.06 & 2.09 & -4.16 & -5.03 & 6.69 & 5.25 & 77.5 & 32.4 & 32.4 & 94.1 \\
        Llama-3.1-8B & 4.69 & 3.25 & 2.06 & -1.94 & -4.06 & 4.69 & 2.38 & 76.5 & 28.6 & 25.5 & 86.7 \\
        \cmidrule(r){1-8}\cmidrule(l){9-12}
        \multicolumn{12}{@{}l}{\emph{Evidence span, task-defined}} \\
        Qwen2.5-7B   & 8.23 & 7.90 & 5.63 & -0.57 & -1.09 & 8.23 & 7.60 & 57.4 & 25.9 & 9.3 & 74.1 \\
        Mistral-7B   & 10.00 & 9.94 & 6.72 & 2.56 & 1.25 & 10.00 & 9.88 & 44.2 & 42.3 & 0.0 & 98.1 \\
        Llama-3.1-8B & 8.62 & 7.62 & 4.50 & 0.50 & -0.69 & 8.62 & 7.75 & 54.9 & 37.3 & 2.0 & 98.0 \\
        \cmidrule(r){1-8}\cmidrule(l){9-12}
        \multicolumn{12}{@{}l}{\emph{Evidence span, response-derived}} \\
        Qwen2.5-7B   & 14.70 & 14.04 & 7.88 & 1.66 & 0.46 & 14.70 & 12.54 & 50.0 & 45.8 & 0.0 & 100.0 \\
        Mistral-7B   & 15.91 & 15.54 & 12.62 & 0.96 & 0.98 & 15.91 & 15.19 & 43.5 & 30.4 & 0.0 & 100.0 \\
        Llama-3.1-8B & 5.06 & 4.84 & 2.01 & -4.62 & -9.28 & 5.06 & 5.23 & 87.5 & 12.5 & 4.2 & 95.8 \\
        \bottomrule
    \end{tabular}
    \par\vspace{2pt}
    \begin{minipage}{0.95\columnwidth}
        \scriptsize
        \raggedright
        \textit{\textbf{Note}.} Let $y_t^\star$ denote the target token at this position, e.g., the token specified by the rule for rule decisions. We measure its local decision advantage using the \textit{target token margin} $m_t=\log p_\theta(y_t^\star\mid P_t)-\max_{v\neq y_t^\star}\log p_\theta(v\mid P_t)$. Here, $m_t>0$ indicates that the target token is top-ranked, while $m_t=0$ marks the decision boundary. Task-defined decisions follow the specified factual requirements, whereas response-derived decisions are identified from facts correctly stated in the models' own generations.
    \end{minipage}
    \vspace{-4pt}
\end{table}

\subsubsection{Recoverability of Failed Decisions}
\label{sec:experiments-3-2}

We have focused on initially successful decisions, where attenuation shows that the local choice depends on sustained role span support. We next examine whether amplifying the same contributions can restore decisions that initially fail. To test the causal effect, we amplify the same contribution at naturally failing decisions and track the task performance.

Considering task-defined cases, we find that amplifying the evidence span contribution restores a substantial subset of naturally failing factual decisions. As shown in Table~\ref{tab:decision_restoration}, the restoration rate reaches 43.7\% at $\beta=2$, accompanied by improved task performance. In contrast, rule failures exhibit much larger initial decision deficits, and their target tokens remain unrecovered even under $16\times$ amplification. These results suggest that whether contribution amplification can reverse a failed decision depends on the initial decision deficit.

We further examine whether contribution amplification can restore factual decisions in genuine hallucinations from RAGTruth. As shown in the real-hallucinated rows of Table~\ref{tab:decision_restoration}, these decisions exhibit larger initial deficits than task-defined evidence decisions. Nevertheless, amplification shifts the decision margin toward the evidence-supported token and improves factual selection. At $\beta=4$, target selection reaches 32.5\%, with 57.5\% of continuations stating the evidence-supported fact, both outperforming random-direction interventions. These results show that evidence span contributions remain effective in genuine hallucinations, although their ability to reverse factual decisions still depends on the decision deficit.

Consequently, these results show that the same local contribution mechanism can extend to naturally failing decisions, including genuine hallucination examples. Whether amplification reverses the choice depends on the balance between the support of role contribution and the decision deficit. This relation further yields a quantitative prediction for failed decisions, which we formalize in Appendix~\ref{app:decision-boundary} and evaluate in Appendix~\ref{app:task_recovery_prediction}. 

\begin{table}[htbp]
    \centering
    \setlength{\abovecaptionskip}{2pt}
    \setlength{\belowcaptionskip}{3pt}
    \caption{\textbf{Restoration at failing decisions}. We restore the contribution vector at the decision, scaled by a weight $\beta$. \textbf{(a)} Restoring the contribution at $\beta{=}1$. \textbf{(b)} Restoration across multiple weights.}
    \label{tab:decision_restoration}
    \footnotesize
    \setlength{\tabcolsep}{2.6pt}
    \begin{tabular}{@{}l cccc @{\hspace{10pt}} l ccccccc@{}}
        \toprule
        & \multicolumn{4}{c}{\textbf{(a) Restoration at $\beta{=}1$}} & \multicolumn{8}{c}{\textbf{(b) Restoration across weights}} \\
        \cmidrule(lr){2-5}\cmidrule(l){6-13}
        & Baseline & Margin & Selection & Task & & & & & & & & \\
        Model & margin & change & (\%) & (\%) & Metric & $\beta{=}0$ & $\beta{=}0.5$ & $\beta{=}1$ & $\beta{=}2$ & $\beta{=}4$ & $\beta{=}8$ & $\beta{=}16$ \\
        \midrule
        \multicolumn{5}{@{}l}{\emph{Rule span}} & & & & & & & & \\
        Qwen2.5-7B   & -14.6 & -0.61 & 0.0 & 0.0 & Selection & 0.0 & 0.0 & 0.0 & 0.0 & 0.0 & 0.0 & 0.0 \\
        Mistral-7B   & -19.0 & +0.06 & 0.0 & 4.8 & Margin & -14.6 & -14.6 & -14.5 & -14.1 & -13.0 & -12.8 & -11.7 \\
        Llama-3.1-8B & -11.3 & +0.00 & 0.0 & 0.0 & Task & 0.0 & 0.0 & 1.3 & 2.5 & 6.3 & 3.8 & 6.3 \\
        \cmidrule(r){1-5}\cmidrule(l){6-13}
        \multicolumn{5}{@{}l}{\emph{Evidence span}, task-defined} & & & & & & & & \\
        Qwen2.5-7B   & -4.0 & +0.89 & 29.5 & 34.1 & Selection & 0.0 & 22.3 & 28.2 & 43.7 & 41.7 & 39.8 & 38.8 \\
        Mistral-7B   & -2.2 & +0.28 & 16.7 & 20.8 & Margin & -3.0 & -2.9 & -2.0 & -1.2 & -0.6 & -1.9 & -2.8 \\
        Llama-3.1-8B & -2.7 & +1.69 & 34.3 & 37.1 & Task & 0.0 & 25.2 & 32.0 & 47.6 & 53.4 & 55.3 & 58.3 \\
        \cmidrule(r){1-5}\cmidrule(l){6-13}
        \multicolumn{5}{@{}l}{\emph{Evidence span}, real-hallucinated} & & & & & & & & \\
        Qwen2.5-7B   & -6.0 & +1.18 & 11.1 & 44.4 & Selection & 0.0 & 8.8 & 10.0 & 17.5 & 32.5 & 33.8 & 32.5 \\
        Mistral-7B   & -5.3 & +1.20 & 11.5 & 42.3 & \hspace{0.8em}\textit{random} & 0.0 & 1.2 & 5.0 & 6.2 & 5.0 & 0.0 & 2.5 \\
        Llama-3.1-8B & -4.5 & +0.96 & 7.4 & 37.0 & Task & 28.8 & 37.5 & 41.2 & 43.8 & 57.5 & 57.5 & 61.2 \\
        All models   & -5.4 & +1.12 & 10.0 & 41.2 & \hspace{0.8em}\textit{random} & 28.8 & 27.5 & 27.5 & 28.8 & 25.0 & 15.0 & 25.0 \\
        \bottomrule
    \end{tabular}
    \par\vspace{2pt}
    \begin{minipage}{0.95\columnwidth}
        \scriptsize
        \raggedright
        \textit{\textbf{Note}.} Margins follow the definition in Table~\ref{tab:decision_attenuation}, and $\Delta m_t=m_t(\beta{=}1)-m_t(\beta{=}0)$; positive values indicate a shift toward the target token. Real-hallucinated decisions are taken from RAGTruth; their \textit{Task} is the share of continuations stating the grounded value at the hallucinated slot. \textit{random} rows replace the contribution vector with a random direction of equal norm.
    \end{minipage}
    \vspace{-4pt}
\end{table}

In summary, we show that dynamic context guides LLM generation through functional role span contributions in both ICL and OOC hallucination. Specifically, rule contributions shape structural and selection choices, whereas evidence contributions shape factual choices. Moreover, effective role influence depends on coherent contribution organization across layers and sufficient support when the key choice is made. The causal interventions further show that a failed choice can be reversed when the corresponding support is sufficient to compensate for the decision deficit.

\section{Conclusion}
In this paper, we offer a common mechanistic view of the distinct outcomes of ICL and OOC hallucination. This view treats the functional roles of dynamic context as the organizing principle for model behavior. We connect internal role representations to their effects on decoding through role span contributions. These contributions causally shape key token choices that determine response structure, content selection, and factual value. Each functional role determines where these contributions must remain effective during generation. We derive a first-order decomposition of role span contributions and identify two necessary conditions for reliable influence. At the decision level, rule following requires sustained support across repeated structural and selection choices, whereas evidence grounding relies on support at a smaller set of decisive factual choices. The contrasting outcomes therefore follow from distinct requirements for sustaining role support across generation.

\subsection*{AI use statement}
In this work, we did not use generative AI tools for tasks requiring mandatory disclosure under the ICLR 2027 AI Policy. Additionally, we used generative AI tools to create or edit software code and to edit the research paper to improve readability. We have reviewed all AI-assisted work. AI-assisted code was verified and tested by the authors, and AI-assisted edits to the manuscript were reviewed by the authors. We take responsibility for the final content of this work, including text, claims or artifacts produced with the aid of generative AI.

\subsection*{Reproducibility statement}
The source code used for our experiments is provided in the supplementary materials and will be released in a public repository. Appendix B provides the complete proofs and derivations of the theoretical results. Appendix C provides the experimental details, including the experimental setting, matched role construction, representation and recovery analyses, and contribution and preference analysis. Appendix D reports additional experimental results.

\bibliography{iclr2027_conference}
\bibliographystyle{iclr2027_conference}

\clearpage
\appendix

\titleformat{\section}
  {\large\bfseries}
  {\thesection.}
  {0.6em}
  {}

\titleformat{\subsection}
  {\normalsize\bfseries}
  {\thesubsection.}
  {0.6em}
  {}

\titlespacing*{\section}
  {0pt}
  {2.0ex plus 0.4ex minus 0.2ex}
  {1.0ex}

\titlespacing*{\subsection}
  {0pt}
  {1.7ex plus 0.4ex minus 0.2ex}
  {0.8ex}

\setlength{\abovedisplayskip}{6pt plus 2pt minus 2pt}
\setlength{\belowdisplayskip}{6pt plus 2pt minus 2pt}
\setlength{\abovedisplayshortskip}{4pt plus 1pt minus 1pt}
\setlength{\belowdisplayshortskip}{4pt plus 1pt minus 1pt}

\noindent{\Large\bfseries Appendix}
\vspace{1.5ex}

\section{Related Work}
\label{app:related-work}

\paragraph{Context-conditioned generation.}

Motivated by the contextual adaptability of LLMs, recent work has extensively studied how information supplied at inference time shapes model behavior. In-context learning provides a prominent setting where models can infer a task from demonstrations and instructions without task-specific training~\citep{gu2023pretraining,ye2023compositional}. Recent studies further show that this adaptation is sensitive to demonstration selection, ordering, and composition~\citep{lu2022ordering,ye2023compositional,peng2024selection,wang2025selection}. Meanwhile, a separate line of work has investigated hallucination in LLM generation, where model outputs contain factually incorrect or unsupported information~\citep{huang2025hallucination,yu2024mechanistic}. Retrieval-augmented generation has become a common approach to mitigating hallucination by incorporating external evidence into generation~\citep{yoran2024robust,asai2024selfrag}. Nevertheless, LLMs may still produce claims that are unsupported by or contradictory to the retrieved content~\citep{niu-etal-2024-ragtruth,wu2024clasheval}. Such departures from the supplied context define the out-of-context (OOC) hallucination setting and expose a gap in understanding contextual faithfulness~\citep{asai2024selfrag,huang2025faithfulness}.

\paragraph{Mechanistic analyses of inference-time behavior.}

Recent mechanistic work extends the study of inference-time behavior from observable outcomes to internal representations. In ICL, some studies identify attention circuits that support in-context computation~\citep{ren2024identifying}, while other work seeks to explicitly characterize in-context task information by recovering task vectors from hidden states~\citep{hendel2023task,todd2024function}. In parallel, mechanistic studies of hallucination associate incorrect or unsupported generation with failures in knowledge retrieval and transfer and with conflicts between parametric knowledge and external evidence~\citep{yu2024mechanistic,xie2024adaptive}. Other studies further localize components involved in retrieving and using contextual information during generation, showing that evidence use can likewise be traced to internal components~\citep{wu2025retrieval,neo2024context}. Despite this progress, the two behaviors are predominantly examined within specific tasks or failure settings. Across both lines, a common concern is which internal representations of context become consequential for generation.

\paragraph{Attributing contextual influence.}

To establish this relevance, existing methods connect model predictions to their internal sources. For instance, input attribution and score decomposition quantify how tokens or activations support a prediction~\citep{sundararajan2017axiomatic}, while circuit analysis traces it to interpretable computational paths~\citep{dunefsky2024transcoders}. However, such local analyses rarely make explicit the function of context in the current task~\citep{yu2024mechanistic}. Despite comparable mechanistic accounts, ICL and OOC hallucination exhibit distinct task-level behaviors. To this end, we analyze the distinct functional roles in the two settings and identify the mechanistic conditions underlying task success and failure from a unified perspective.

\section{Proofs and Derivations}
\label{app:proofs}

\subsection{Attention-Mediated Role Span Transport}
\label{app:attention-contribution}

We first formalize why a role-induced difference can enter a new
token position only through attention.

\begin{applemma}[Position locality]
\label{lem:position-locality}
In a decoder-only Transformer, normalization and MLP modules act
independently at each token position, whereas the attention update at
position \(i\) aggregates states from the source positions visible to
\(i\).
\end{applemma}

\begin{proof}
Normalization and MLP modules apply the same map independently at
each position. By contrast,
Eq.~\eqref{eq:attn-output-source-decomposition} writes the attention
update at \(i\) as a sum over its visible source positions. Position-wise
transformations of queries and keys do not alter this distinction.
\end{proof}

Fix a decoding step \(t\) and write
\(P=P_t^f(y)\), \(\widetilde P=\widetilde P_t^f(y)\), and
\(\mathcal S_f\) for their aligned role span positions. Define
\[
d_\ell(i)=h_\ell(P,i)-h_\ell(\widetilde P,i),
\qquad
\mathcal C_\ell=\{s:d_\ell(s)\neq 0\},
\]
and let \(\mathcal V(i)\) denote the source positions visible to \(i\).
For head \(a\) and source position \(s\in\mathcal V(i)\), write
\[
g_{\ell,i,s}^{(a)}(P)
=
\alpha_{i\leftarrow s}^{(a)}(P)
V_\ell^{(a)}(P,s)W_O^{(a)},
\qquad
\Delta g_{\ell,i,s}^{(a)}
=
g_{\ell,i,s}^{(a)}(P)
-
g_{\ell,i,s}^{(a)}(\widetilde P).
\]
We decompose the source-indexed differences as
\[
\begin{aligned}
\Delta B_{\ell,i}^f
&=
\sum_a
\sum_{s\in\mathcal V(i)\cap\mathcal S_f}
\Delta g_{\ell,i,s}^{(a)},\\
\Delta I_{\ell,i}
&=
\sum_a
\sum_{s\in\mathcal V(i)\cap
(\mathcal C_{\ell-1}\setminus\mathcal S_f)}
\Delta g_{\ell,i,s}^{(a)},\\
\Delta W_{\ell,i}
&=
\sum_a
\sum_{s\in\mathcal V(i)\setminus
(\mathcal S_f\cup\mathcal C_{\ell-1})}
\Delta g_{\ell,i,s}^{(a)}.
\end{aligned}
\]

\begin{appproposition}[Attention-mediated role span transport]
\label{prop:attention-mediated-transport}
Suppose that \(P\) and \(\widetilde P\) differ only at the aligned role
span positions \(\mathcal S_f\). For every
\(i\notin\mathcal S_f\), \(d_0(i)=0\), and at each layer
\(\ell\geq 1\),
\begin{equation}
\Delta\operatorname{Attn}_\ell(i)
=
\Delta B_{\ell,i}^f
+
\Delta I_{\ell,i}
+
\Delta W_{\ell,i},
\label{eq:attention-difference-partition}
\end{equation}
where
\[
\Delta\operatorname{Attn}_\ell(i)
=
\operatorname{Attn}^{\mathrm{out}}_\ell(P,i)
-
\operatorname{Attn}^{\mathrm{out}}_\ell(\widetilde P,i).
\]
Moreover,
\[
d_{\ell-1}(i)=0
\quad\text{and}\quad
\Delta\operatorname{Attn}_\ell(i)=0
\quad\Longrightarrow\quad
d_\ell(i)=0.
\]
Consequently, if
\(\ell^\star=\min\{\ell:d_\ell(i)\neq 0\}\) exists, then
\(\Delta\operatorname{Attn}_{\ell^\star}(i)\neq 0\).
\end{appproposition}

\begin{proof}
Since the matched inputs differ only at \(\mathcal S_f\), their
embedding representations coincide at every
\(i\notin\mathcal S_f\). Hence, \(d_0(i)=0\).

We next decompose the attention difference. By
Eq.~\eqref{eq:attn-output-source-decomposition},
\[
\Delta\operatorname{Attn}_\ell(i)
=
\sum_a\sum_{s\in\mathcal V(i)}
\Delta g_{\ell,i,s}^{(a)}.
\]
The visible source positions admit the disjoint partition
\[
\mathcal V(i)
=
\bigl(\mathcal V(i)\cap\mathcal S_f\bigr)
\mathbin{\dot\cup}
\bigl(\mathcal V(i)\cap
(\mathcal C_{\ell-1}\setminus\mathcal S_f)\bigr)
\mathbin{\dot\cup}
\bigl(\mathcal V(i)\setminus
(\mathcal S_f\cup\mathcal C_{\ell-1})\bigr).
\]
Summing over these three sets gives
Eq.~\eqref{eq:attention-difference-partition}. The first term is the
direct role span contribution defined in
Eq.~\eqref{eq:role span-attention-contribution}; the second reads from
non-role positions whose residual states already differ.

For a source position in the third set,
\(d_{\ell-1}(s)=0\), and hence its value representation is identical
under the two computations:
\[
V_\ell^{(a)}(P,s)
=
V_\ell^{(a)}(\widetilde P,s).
\]
Thus, \(\Delta W_{\ell,i}\) contains the remaining changes in
attention weights over otherwise unchanged source values. This
definition covers any such reweighting without assigning it to a
particular query, key, or softmax component.

We next show how the difference propagates through the block.
If \(d_{\ell-1}(i)=0\) and
\(\Delta\operatorname{Attn}_\ell(i)=0\), the two computations provide
the same local state and the same attention update at position \(i\).
By Lemma~\ref{lem:position-locality}, all remaining operations at
that position therefore receive identical inputs, which gives
\(d_\ell(i)=0\).

Now let \(\ell^\star\) be the first layer such that
\(d_{\ell^\star}(i)\neq 0\). Then
\(d_{\ell^\star-1}(i)=0\). If the attention updates also agreed at
layer \(\ell^\star\), the previous implication would yield
\(d_{\ell^\star}(i)=0\), a contradiction. Therefore,
\(\Delta\operatorname{Attn}_{\ell^\star}(i)\neq 0\).
\end{proof}

Proposition~\ref{prop:attention-mediated-transport} establishes that
attention is the direct channel through which a role-induced
difference enters a new token position. Among the resulting terms,
\(\Delta B_{\ell,i}^f\) is the direct source-indexed role span pathway
used in Section~\ref{sec:role span-contributions}, while
\(\Delta I_{\ell,i}\) and \(\Delta W_{\ell,i}\) are treated in the
remainder analysis of Appendix~\ref{app:remainder}.

\subsection{First-Order Role Span Score Decomposition}
\label{app:local-score-decomposition}

We now make the coordinate dependence in
Assumption~\ref{ass:local-score-regularity} explicit. Fix a decoding
step \(t\) and write
\[
\widetilde b
=
\left(
B_{\ell,i}^f(\widetilde P_t^f(y))
\right)_{(\ell,i)\in\mathcal B_{f,t}},
\quad
b^f
=
\left(
B_{\ell,i}^f(P_t^f(y))
\right)_{(\ell,i)\in\mathcal B_{f,t}},
\]
with
\[
\Delta b
=
b^f-\widetilde b
=
\Delta B_{\mathcal B,t}^f(y).
\]
We equip this stacked coordinate space with the Euclidean inner product and norm.
\paragraph{Substitution map.}
For \(b=(b_{\ell,i})_{(\ell,i)\in\mathcal B_{f,t}}\), let \(\Psi_t(b)\) denote the score obtained from \(\widetilde P_t^f(y)\) by processing the model in forward order and replacing each selected role span attention contribution with \(b_{\ell,i}\) before it enters the residual stream. Specifically, the term recomputed from the current modified state is removed and replaced by \(b_{\ell,i}\). All replacements within a layer are applied before advancing to the next layer, and the remaining quantities are computed
from the resulting states.

\begin{applemma}[Exact substitution identity]
\label{lem:exact-substitution-identity}
The substitution map is well defined and satisfies
\[
\Psi_t(\widetilde b)
=
\psi_t(\widetilde P_t^f(y)).
\]
Consequently, with
\[
E_t^\perp
=
\psi_t(P_t^f(y))-\Psi_t(b^f),
\]
the stepwise score shift admits the exact identity
\begin{equation}
\delta_t^f(y)
=
\Psi_t(b^f)-\Psi_t(\widetilde b)+E_t^\perp.
\label{eq:exact-substitution-identity}
\end{equation}
\end{applemma}

\begin{proof}
The forward ordering uniquely determines every modified residual state, so \(\Psi_t(b)\) is well defined for any collection \(b\).

We next establish the endpoint identity. Set \(b=\widetilde b\). At the input layer, the modified computation and the computation under \(\widetilde P_t^f(y)\) have identical states. Suppose that their states agree through layer \(\ell-1\). Their unmodified attention terms at layer \(\ell\) are then identical, and each selected role span term is replaced by its value \(B_{\ell,i}^f(\widetilde P_t^f(y))\). Hence, the complete attention updates and the resulting residual states also agree at layer \(\ell\). Induction over the layers gives \(\Psi_t(\widetilde b) = \psi_t(\widetilde P_t^f(y)).\)

Finally, adding and subtracting \(\Psi_t(b^f)\) in the definition of
\(\delta_t^f(y)\) gives
\[
\begin{aligned}
\delta_t^f(y)
&=
\psi_t(P_t^f(y))
-
\psi_t(\widetilde P_t^f(y))
\\
&=
\Psi_t(b^f)-\Psi_t(\widetilde b)
+
\bigl[
\psi_t(P_t^f(y))-\Psi_t(b^f)
\bigr],
\end{aligned}
\]
which is Eq.~\eqref{eq:exact-substitution-identity}.
\end{proof}

\begin{proof}[
Proof of Theorem~\ref{theorem:First-order role span score decomposition}]
By Lemma~\ref{lem:exact-substitution-identity}, it remains to expand
\(\Psi_t(b^f)-\Psi_t(\widetilde b)\). Let
\[
\gamma(\tau)=\widetilde b+\tau\Delta b,
\qquad
\tau\in[0,1].
\]
Then we have
\[
\Psi_t(b^f)-\Psi_t(\widetilde b)
=
\int_0^1
\left\langle
\nabla\Psi_t(\gamma(\tau)),
\Delta b
\right\rangle
\,d\tau.
\]
Adding and subtracting \(\nabla\Psi_t(\widetilde b)\) gives
\[
\Psi_t(b^f)-\Psi_t(\widetilde b)
=
\left\langle
\nabla\Psi_t(\widetilde b),
\Delta b
\right\rangle
+
r_t^{\mathrm{curv}},
\]
where
\[
r_t^{\mathrm{curv}}
=
\int_0^1
\left\langle
\nabla\Psi_t(\gamma(\tau))
-
\nabla\Psi_t(\widetilde b),
\Delta b
\right\rangle
\,d\tau.
\]

Since \(\Psi_t\) has an \(L_t\)-Lipschitz gradient on the segment
between \(\widetilde b\) and \(b^f\),
\[
\begin{aligned}
|r_t^{\mathrm{curv}}|
&\le
\int_0^1
\left\|
\nabla\Psi_t(\gamma(\tau))
-
\nabla\Psi_t(\widetilde b)
\right\|
\|\Delta b\|
\,d\tau
\\
&\le
\int_0^1
L_t\tau\|\Delta b\|^2
\,d\tau
=
\frac{L_t}{2}\|\Delta b\|^2.
\end{aligned}
\]

Moreover, the stacked inner product decomposes as
\[
\left\langle
\nabla\Psi_t(\widetilde b),
\Delta b
\right\rangle
=
\sum_{(\ell,i)\in\mathcal B_{f,t}}
\left\langle
\nabla_{b_{\ell,i}}\Psi_t(\widetilde b),
\Delta B_{\ell,i,t}^f(y)
\right\rangle.
\]
Combining this equality with
Eq.~\eqref{eq:exact-substitution-identity} yields
Eq.~\eqref{eq:first-order-role span-score-shift} with
\[
R_t=r_t^{\mathrm{curv}}+E_t^\perp.
\]
Finally,
\[
|R_t|
\le
|r_t^{\mathrm{curv}}|+|E_t^\perp|
\le
\frac{L_t}{2}
\left\|
\Delta B_{\mathcal B,t}^f(y)
\right\|^2
+
|E_t^\perp|,
\]
which gives
Eq.~\eqref{eq:first-order-role span-score-shift-remainder}.
\end{proof}
The term \(r_t^{\mathrm{curv}}\) is the Taylor remainder within the selected coordinate space, whereas \(E_t^\perp\) contains the score change not reproduced by these coordinates. We characterize the latter terms in Appendix~\ref{app:remainder}.

\subsection{Remainder Characterization}
\label{app:remainder}

By Theorem~\ref{theorem:First-order role span score decomposition}, the remainder separates as  \(R_t = r_t^{\mathrm{curv}}+E_t^\perp\), where \(|R_t|
\le
\frac{L_t}{2}
\left\|
\Delta B_{\mathcal B,t}^f(y)
\right\|^2
+
|E_t^\perp|.\)

Here, \(r_t^{\mathrm{curv}}\) is the curvature error introduced by linearizing \(\Psi_t\) over the selected coordinates, whereas \(E_t^\perp=\psi_t(P_t^f(y))-\Psi_t(b^f)\) measures the score change not reproduced by these substitutions. Relative to the decomposition in Appendix~\ref{app:attention-contribution}, the substitution gap may include omitted direct coordinates and portions of \(\Delta I_{\ell,i}\) and \(\Delta W_{\ell,i}\) that are not recovered by the modified computation.

Notably, this characterization does not assume that \(R_t\) is small. In Appendix~\ref{app:role span-attention-pathway}, we compare the direct role span attention pathway with alternative transport terms and MLP updates, and show that attention provides the dominant route for role information to reach the relevant token decisions.

\subsection{Decision-Level Recovery Prediction}
\label{app:decision-boundary}

We finally relate the role span contribution to the local decision
boundary considered in Section~\ref{sec:Role Span Contributions to LLM Decision}. For a naturally failing decision,
let \(y_t^\star\) denote the target token and let
\[
c_t
=
\arg\max_{v\neq y_t^\star}
\log p_\theta(v\mid P_t;\beta=0)
\]
be its strongest competitor before amplification. Following
Section~\ref{sec:experiments-3-2}, let \(\beta\) parameterize amplification of the same role span contribution \(B_{\ell,i}^f(P)\) at the decision position, with \(\beta=0\) denoting the original decision. For the fixed competitor \(c_t\), write
\[
m_t(\beta;c_t)
=
\log p_\theta(y_t^\star\mid P_t;\beta)
-
\log p_\theta(c_t\mid P_t;\beta),
\]
so that \(m_t(0;c_t)=m_t(0)<0\). We define the signed support supplied
along the amplification direction as
\[
S_t
=
\left.
\frac{d}{d\beta}m_t(\beta;c_t)
\right|_{\beta=0}.
\]

\begin{appproposition}[Decision-boundary crossing]
\label{prop:decision-boundary-crossing}
Suppose that the derivative of \(m_t(\beta;c_t)\) is
\(L\)-Lipschitz along the local amplification path. Then
\begin{equation}
\left|
m_t(\beta;c_t)-m_t(0)-\beta S_t
\right|
\le
\frac{L}{2}\beta^2.
\label{eq:decision-margin-expansion}
\end{equation}
For \(m_t(0)<0\) and \(S_t>0\), the corresponding first-order
prediction of the amplification required to reach the decision
boundary is
\begin{equation}
\widehat{\beta}_t
=
-\frac{m_t(0)}{S_t}.
\label{eq:predicted-crossing-strength}
\end{equation}
Moreover, if \(c_t\) remains the strongest non-target token along the
local path, then any \(\beta>0\) satisfying
\begin{equation}
\beta S_t
>
-m_t(0)
+
\frac{L}{2}\beta^2
\label{eq:sufficient-decision-crossing}
\end{equation}
guarantees \(m_t(\beta)>0\).
\end{appproposition}

\begin{proof}
A first-order expansion of the fixed-competitor margin at
\(\beta=0\) gives
\[
m_t(\beta;c_t)
=
m_t(0)
+
\beta S_t
+
\varepsilon_t(\beta),
\]
where the Lipschitz condition yields
\(
|\varepsilon_t(\beta)|
\le
L\beta^2/2
\).
This proves Eq.~\eqref{eq:decision-margin-expansion}.

Ignoring the quadratic term, the local boundary is reached when
\(m_t(0)+\beta S_t=0\). For \(S_t>0\), solving this equality gives
Eq.~\eqref{eq:predicted-crossing-strength}. Thus, a larger initial
deficit \(-m_t(0)\) requires stronger amplification, whereas larger
positive signed support \(S_t\) reduces the required strength.

For the finite-\(\beta\) statement,
Eq.~\eqref{eq:decision-margin-expansion} gives the lower bound
\[
m_t(\beta;c_t)
\ge
m_t(0)
+
\beta S_t
-
\frac{L}{2}\beta^2.
\]
Hence, Eq.~\eqref{eq:sufficient-decision-crossing} implies
\(m_t(\beta;c_t)>0\). If \(c_t\) remains the strongest competitor,
this pairwise margin equals the target-token margin, and therefore
\(m_t(\beta)>0\).
\end{proof}

Consequently, the first-order decision formulation estimates the amplification required to recover a failed decision. Appendix~\ref{app:task_recovery_prediction} further examines the extension of this prediction to task-level recovery.

\section{Experimental Details}
\label{app:experimental-details}

\paragraph{Experimental setting.}
For the rule role, we derive a controlled benchmark from the task patterns
in CL-bench \citep{dou2026clbench}, covering format, selection, and
include or omit instructions. For the evidence role, we use the training
split of RAGTruth \citep{niu-etal-2024-ragtruth}. We evaluate Qwen2.5-7B, Mistral-7B, and Llama-3.1-8B. Free generations use greedy decoding, whereas fixed response trajectories are evaluated with teacher forcing.

\subsection{Matched Role Construction}
\label{app:matched-role-construction}

To isolate functional role $f$, we compare two prompts that share the query and all non-role content but differ in the role-bearing span. The matched counterfactual changes only this span while preserving the remainder of the prompt. Figure~\ref{fig:matched-roles} illustrates the matched construction. To ensure alignment, we locate the edited span separately in each tokenized prompt and, for a fixed response trajectory \(y\), use the same response prefix \(y_{<t}\) under both conditions, with comparisons performed at the same decoding step.

\subsection{Representation and Recovery Analyses}
\label{app:representation-recovery}

\paragraph{Subspace concentration.}
We extract \(\Delta h_\ell^f\) at the final prompt position from every
decoder layer. For each model and role, we fit a rank-\(10\) subspace
on one half of the samples and evaluate it on the other, then exchange
the two splits. We use the matched role comparisons defined in Appendix \ref{app:matched-role-construction}. Specifically, the held-out concentration is computed as \(\mathcal{E}_{\ell,i}^{f}=\frac{\|\Pi_{U_\ell}\Delta h_\ell^{f}(i)\|_2^2}{\|\Delta h_\ell^{f}(i)\|_2^2}\), where \(\Pi_{U_\ell}\) projects onto the fitted rank-10 subspace \(U_\ell\). The random-subspace baseline is \(\mathcal{E}_{\mathrm{rand}}=10/H\), where \(H\) is the hidden state dimension.

\paragraph{Output alignment and behavior recovery.}
We assess output alignment by applying the final RMS normalization to the two residual states in each matched pair and then projecting their difference onto the unembedding direction of the token selected under \(C_f\). We report the resulting logit shift and cosine alignment, with a matched non-role perturbation as the control. Separately, we evaluate behavior recovery by adding \(\Delta h_\ell^f\) to the counterfactual state at the final prompt position and recording whether the next-token choice returns to that under \(C_f\). 

\subsection{Contribution and Preference Analysis}
\label{app:contribution-preference}
\paragraph{Candidate comparisons.}
Each preference analysis combines a matched context pair with a
minimal candidate pair. Both candidates are scored with teacher forcing, and \(M_f\) measures the resulting change in relative support. Figure~\ref{fig:matched-preference-example}
illustrates the design.

\begin{figure}[htbp]
\centering
\small
\setlength{\tabcolsep}{5pt}
\renewcommand{\arraystretch}{1.18}
\begin{tabularx}{\linewidth}{
  @{}
  >{\raggedright\arraybackslash}p{0.13\linewidth}
  >{\raggedright\arraybackslash}X
  >{\raggedright\arraybackslash}X
  @{}
}
\toprule
& \textbf{Matched contexts} & \textbf{Matched candidates} \\
\midrule
\textbf{Rule}
& \(C_r\): JSON-output rule\par
  \(\widetilde C_r\): Markdown-list rule
& \(y^+\): \{"\texttt{name}": "Alpha", "value": "120"\}\par
  \(y^-\): \{"\texttt{label}": "Alpha", "value": "120"\} \\
\addlinespace[4pt]
\textbf{Evidence}
& \(C_e\): Evidence supporting \texttt{2024}\par
  \(\widetilde C_e\): Matched context, e.g., evidence supporting \texttt{2029} 
& \(y^+\): The value is \texttt{2024}\par
  \(y^-\): The value is \texttt{2029} \\
\bottomrule
\end{tabularx}
\vspace{3pt}
\[
\begin{aligned}
\Phi_f(y) &= \log p_\theta(y\mid C_f \oplus x)-\log p_\theta(y\mid \widetilde C_f \oplus x),
\quad M_f = \Phi_f(y^+)-\Phi_f(y^-).
\end{aligned}
\]
\caption{\textbf{Matched preference construction}. Each context pair differs only in the role-bearing span, while each candidate pair differs only in the choice specified by that role.}
\label{fig:matched-preference-example}
\end{figure}

\section{Additional Experiments}

\subsection{Additional Model Results for Contribution Organization}
\label{app:contribution-organization}

\begin{figure}[htbp]
  \centering
  \includegraphics[
    width=\columnwidth,
    keepaspectratio
  ]{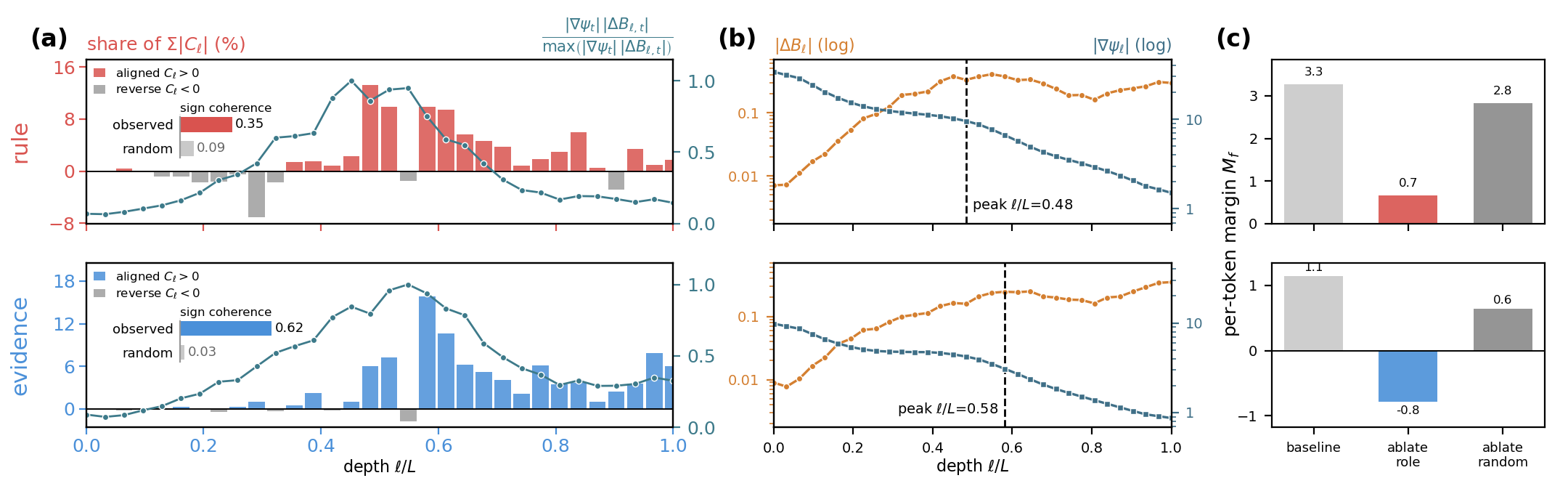}
  \caption{\textbf{Organization of contributions for Mistral-7B}.
  \textbf{(a)} Signed layerwise contributions and sign coherence;
  \textbf{(b)} layerwise profiles of
  \( |\Delta B_{\ell,t}| \) and \( |\nabla \psi_t| \);
  and \textbf{(c)} preference margin \(M_f\) under the baseline,
  role span pathway ablation, and random ablation.}
  \label{fig:mistral_contribution_organization}
\end{figure}

\begin{figure}[htbp]
  \centering
  \includegraphics[
    width=\columnwidth,
    keepaspectratio
  ]{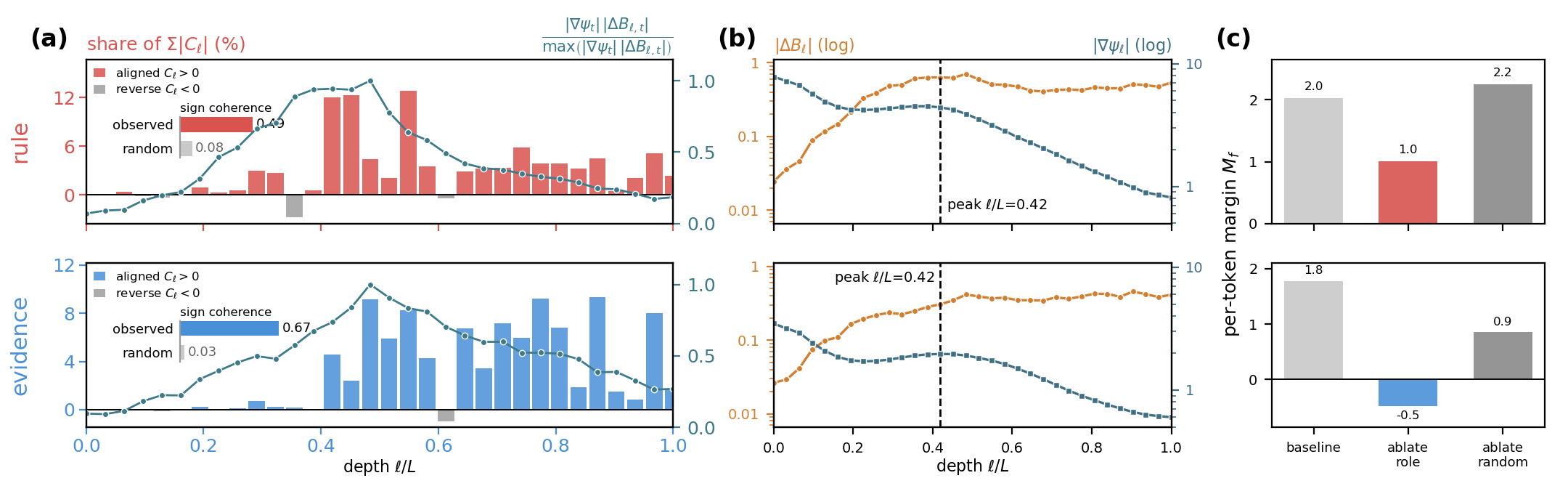}
  \caption{\textbf{Organization of contributions for Llama-3.1-8B}.
  \textbf{(a)} Signed layerwise contributions and sign coherence;
  \textbf{(b)} layerwise profiles of
  \( |\Delta B_{\ell,t}| \) and \( |\nabla \psi_t| \);
  and \textbf{(c)} preference margin \(M_f\) under the baseline,
  role span pathway ablation, and random ablation.}
  \label{fig:llama_contribution_organization}
\end{figure}

Figures~\ref{fig:mistral_contribution_organization}
and~\ref{fig:llama_contribution_organization} report the corresponding
analyses for Mistral-7B and Llama-3.1-8B, both of which exhibit the same
overall pattern. Contributions remain concentrated within a limited
range of layers, the two magnitude factors vary in opposite directions
across depth, and role span pathway ablation reduces \(M_f\) more than
random ablation. The contribution organization identified in the main
analysis is therefore consistent across models.

\begin{table}[htbp]
  \centering 
  \setlength{\abovecaptionskip}{2pt}
  \setlength{\belowcaptionskip}{3pt}
  \setlength{\aboverulesep}{0.3ex}
  \setlength{\belowrulesep}{0.3ex}
  \caption{\textbf{Correspondence between first-order contributions and exact score shifts}.}
  \label{tab:first_order_corr}
  \begin{tabular}{@{}l@{\hspace{8pt}}c@{\hspace{6pt}}c@{\hspace{10pt}}c@{\hspace{6pt}}c@{}}
  \toprule
  & \multicolumn{2}{c}{\makecell[t]{Rule span}}
  & \multicolumn{2}{c}{\makecell[t]{Evidence span}} \\
  \cmidrule(lr){2-3}\cmidrule(l){4-5}
  Model
  & \makecell[b]{Corr.} & \makecell[b]{Sign (\%)}
  & \makecell[b]{Corr.} & \makecell[b]{Sign (\%)} \\
  \midrule
  Qwen2.5-7B & 0.93 & 77.1 & 0.86 & 95.0 \\
  Mistral-7B & 0.94 & 87.6 & 0.88 & 97.0 \\
  Llama-3.1-8B & 0.92 & 91.4 & 0.85 & 94.0 \\
  \bottomrule
  \end{tabular}
  \par\smallskip
  \parbox{\textwidth}{\footnotesize\textit{\textbf{Note}.} \textit{Corr.} is the median Spearman correlation between the magnitudes of the summed first-order contributions and the exact stepwise score shifts. \textit{Sign} is the fraction of responses whose accumulated contribution matches the sign of the exact trajectory shift.}
  \end{table}

\subsection{First-Order Correspondence}
\label{app:first_order_corr}
Here we examine how the first-order contributions correspond to the exact role effect. Within each response, the magnitudes of the summed first-order contributions track the exact stepwise score shifts, with median Spearman correlations of 0.85--0.94 across models. The sign of the accumulated contribution also matches the exact trajectory shift for most responses.

\subsection{Role Span Attention Pathway}
\label{app:role span-attention-pathway}

Appendix~\ref{app:attention-contribution} establishes that attention is the direct channel when a role-induced difference enters a new token position. To examine how much of the realized role effect this pathway carries, we study its recoverability across attention terms and its decision effects. Specifically, for each matched context pair, we fix the exact trajectory score effect \(\Phi_f\), and substitute one term at a time. We consider the direct role term \(\Delta B\), attention reweighting \(\Delta W\), and a matched non-role term. 

Table~\ref{tab:attention-pathway-recovery} shows that the direct term recovers \(73.5\%\)--\(93.9\%\) of the exact role effect across models. In contrast, recovery from attention reweighting and matched non-role positions remains near zero. Together with the sustained attenuation results in Section~\ref{sec:experiments-3-1}, these findings show that the direct role span attention term carries a substantial and behaviorally consequential part of the role effect.

Since it is essential for the role effect to become available at the decision, we next attenuate the carried state at three depths. Table~\ref{tab:module_specific_decision_effects}(a) shows that attenuation at \(0.25L\) has little effect in any setting. By \(0.50L\), the dependence begins to emerge, and at \(0.75L\), the margin drop is positive for all settings. Thus, the role effect becomes relevant to the decision only after sufficient depth. Table~\ref{tab:module_specific_decision_effects}(b) shows that attenuating either update family reduces the target margin. Across comparisons, the attention intervention produces a larger effect than the MLP intervention. 

Taken together, these results establish the empirical importance of the role span attention pathway. The direct attention term recovers a substantial part of the role effect, and this effect becomes increasingly available from the carried state with depth. Combined with the consistently larger intervention effect of attention than MLP, this pattern supports attention as the dominant route through which role information reaches the relevant token decisions.

\begin{table*}[htbp]
    \centering
    \setlength{\abovecaptionskip}{2pt}
    \setlength{\belowcaptionskip}{3pt}
    \caption{\textbf{Direct recovery of role effects.}
    Median normalized recovery (\%) after substituting the direct role term
    ($\Delta B$), attention reweighting ($\Delta W$), or matched non-role positions.}
    \label{tab:attention-pathway-recovery}
    \small
    \setlength{\tabcolsep}{6pt}
    \begin{tabular}{@{}l rrr rrr@{}}
        \toprule
        & \multicolumn{3}{c}{\textbf{Rule}} & \multicolumn{3}{c}{\textbf{Evidence}} \\
        \cmidrule(lr){2-4}\cmidrule(l){5-7}
        Model & Direct $\Delta B$ & Reweighting $\Delta W$ & Non-role & Direct $\Delta B$ & Reweighting $\Delta W$ & Non-role \\
        \midrule
        Qwen2.5-7B & 73.5 & 0.0 & 0.1        & 85.9 & 1.9 & 1.4 \\
        Mistral-7B & 88.3 & 0.6 & $-0.9$     & 92.0 & 1.9 & $-0.7$ \\
        Llama-3.1-8B & 87.7 & 0.8 & $-0.2$     & 93.9 & 0.4 & 0.8 \\
        \bottomrule
    \end{tabular}
\end{table*}
\begin{table*}[htbp]
    \centering
    \setlength{\abovecaptionskip}{2pt}
    \setlength{\belowcaptionskip}{3pt}
    \caption{\textbf{Accumulated carry and update-family effects at role-sensitive decisions.} \textbf{(a)} The carried state entering the layer at three fractional depths. \textbf{(b)} Each update family at all layers.}
    \label{tab:module_specific_decision_effects}
    \setlength{\tabcolsep}{3pt}
    \resizebox{\linewidth}{!}{%
    \begin{tabular}{@{}lrrr@{\hspace{12pt}}rr@{}}
        \toprule
        & \multicolumn{3}{c}{\textbf{(a) Carry drop by depth}} & \multicolumn{2}{c}{\textbf{(b) Update families}} \\
        \cmidrule(lr){2-4}\cmidrule(l){5-6}
        Model & \multicolumn{1}{c}{$\ell/L{=}0.25$} & \multicolumn{1}{c}{$0.5$} & \multicolumn{1}{c}{$0.75$} & \multicolumn{1}{c}{Attention} & \multicolumn{1}{c}{MLP} \\
        \midrule
        \multicolumn{6}{@{}l}{\emph{Rule}} \\
        \hspace{0.8em}Qwen2.5-7B & \mbox{$-0.01$ {\scriptsize[$-0.13$, 0.12]}} & \mbox{$-0.22$ {\scriptsize[$-0.64$, 0.14]}} & \mbox{11.25 {\scriptsize[7.18, 16.11]}} & \mbox{30.79 {\scriptsize[24.51, 37.68]}} & \mbox{26.14 {\scriptsize[21.55, 31.06]}} \\
        \hspace{0.8em}Mistral-7B & \mbox{0.05 {\scriptsize[$-0.01$, 0.11]}} & \mbox{0.77 {\scriptsize[0.15, 1.42]}} & \mbox{12.56 {\scriptsize[9.76, 15.32]}} & \mbox{15.64 {\scriptsize[12.20, 19.21]}} & \mbox{13.40 {\scriptsize[10.84, 15.97]}} \\
        \hspace{0.8em}Llama-3.1-8B & \mbox{$-0.04$ {\scriptsize[$-0.10$, 0.03]}} & \mbox{3.25 {\scriptsize[2.10, 4.55]}} & \mbox{9.11 {\scriptsize[8.20, 10.00]}} & \mbox{13.74 {\scriptsize[11.59, 15.96]}} & \mbox{8.77 {\scriptsize[7.71, 9.68]}} \\
        \midrule
        \multicolumn{6}{@{}l}{\emph{Evidence}} \\
        \hspace{0.8em}Qwen2.5-7B & \mbox{0.02 {\scriptsize[$-0.24$, 0.28]}} & \mbox{0.23 {\scriptsize[$-0.11$, 0.61]}} & \mbox{4.25 {\scriptsize[3.05, 5.59]}} & \mbox{8.76 {\scriptsize[6.76, 11.01]}} & \mbox{7.18 {\scriptsize[5.43, 9.02]}} \\
        \hspace{0.8em}Mistral-7B & \mbox{$-0.04$ {\scriptsize[$-0.17$, 0.07]}} & \mbox{1.10 {\scriptsize[0.68, 1.53]}} & \mbox{5.00 {\scriptsize[3.83, 6.19]}} & \mbox{8.70 {\scriptsize[6.85, 10.83]}} & \mbox{7.81 {\scriptsize[6.37, 9.27]}} \\
        \hspace{0.8em}Llama-3.1-8B & \mbox{0.20 {\scriptsize[$-0.06$, 0.41]}} & \mbox{2.61 {\scriptsize[1.82, 3.46]}} & \mbox{4.68 {\scriptsize[3.61, 5.76]}} & \mbox{7.90 {\scriptsize[6.22, 9.70]}} & \mbox{4.01 {\scriptsize[2.62, 5.54]}} \\
        \bottomrule
    \end{tabular}%
    }
    \par\smallskip
    \parbox{\textwidth}{\footnotesize\textit{Note.} Drops are computed as matched control minus targeted intervention, so larger values indicate stronger dependence.}
\end{table*}

\subsection{Rule Contributions in Few-Shot ICL}
\label{app:demonstration-icl}

Our main experiments place each rule in an explicit prompt span, allowing us to isolate its effect on generation. To test whether the same account extends to few-shot ICL, we use prompts in which labeled examples implicitly specify the rule. The model must infer the underlying task mapping from these examples. We then test whether the inferred mapping makes a measurable causal contribution to the decision and identify where this contribution is realized within the examples.

We use the SST-2 sentiment task from the Stanford Sentiment Treebank \citep{socher2013recursive} and the six-class TREC question classification task \citep{li2002learning}. To isolate the effect of the inferred mapping, we preserve the demonstrations, their order, the query, and the prompt format across each prompt pair, varying only the assignment of labels to classes. We use the demonstration blocks as the rule positions and matched positions outside them as controls.

\begin{table}[htbp]
    \centering
    \setlength{\abovecaptionskip}{2pt}
    \setlength{\belowcaptionskip}{3pt}
    \caption{\textbf{Role effects for rules induced from demonstrations.} The three column groups report rule induction, contribution alignment, and the effect of targeted attenuation.}
    \label{tab:demo_icl_main}
    \small
    \setlength{\tabcolsep}{2.5pt}
    \begin{tabular}{@{}lccc@{\hspace{9pt}}cc@{\hspace{9pt}}cc@{}}
        \toprule
        & \multicolumn{3}{c}{\textbf{(a) Rule induction}} & \multicolumn{2}{c}{\textbf{(b) Contribution}} & \multicolumn{2}{c}{\textbf{(c) Decision effect}} \\
        \cmidrule(lr){2-4}\cmidrule(lr){5-6}\cmidrule(l){7-8}
        Model & \shortstack{Task acc.\ (\%)\\{\scriptsize orig./perm.}} & \shortstack{Mapping\\{\scriptsize follow (\%)}} & \shortstack{No-demo/\\{\scriptsize broken (\%)}} & \shortstack{Positive\\{\scriptsize margin (\%)}} & \shortstack{Recovery (\%)\\{\scriptsize demo/control}} & \shortstack{Margin drop\\{\scriptsize[95\% CI]}} & \shortstack{Accuracy drop\\{\scriptsize[95\% CI] (pp)}} \\
        \midrule
        \multicolumn{8}{@{}l}{\emph{SST-2}} \\
        \hspace{0.8em}Qwen2.5-7B & 77.1 / 86.7 & 84.8 & 46.7 / 54.3 & 81.9 & 99.4 / 0.1 & 2.27 {\scriptsize[1.7, 2.8]} & 25.7 {\scriptsize[15.2, 35.2]} \\
        \hspace{0.8em}Mistral-7B & 82.9 / 82.9 & 82.9 & 49.5 / 48.6 & 81.9 & 99.8 / 0.0 & 1.24 {\scriptsize[0.8, 1.7]} & 20.0 {\scriptsize[7.6, 31.4]} \\
        \hspace{0.8em}Llama-3.1-8B & 73.3 / 77.1 & 79.0 & 41.0 / 41.9 & 77.1 & 96.2 / $-1.0$ & 0.63 {\scriptsize[0.3, 0.9]} & 15.2 {\scriptsize[6.7, 23.8]} \\
        \midrule
        \multicolumn{8}{@{}l}{\emph{TREC}} \\
        \hspace{0.8em}Qwen2.5-7B & 42.9 / 46.7 & 65.7 & 20.0 / 25.7 & 76.2 & 99.4 / $-0.1$ & 1.04 {\scriptsize[0.4, 1.7]} & 8.6 {\scriptsize[$-1.9$, 20.0]} \\
        \hspace{0.8em}Mistral-7B & 52.4 / 47.6 & 66.7 & 14.3 / 21.9 & 74.3 & 98.5 / $-0.5$ & 1.37 {\scriptsize[0.9, 1.9]} & 28.6 {\scriptsize[18.1, 39.0]} \\
        \hspace{0.8em}Llama-3.1-8B & 35.2 / 37.1 & 73.3 & 14.3 / 17.1 & 66.7 & 97.0 / $-0.7$ & 0.69 {\scriptsize[0.2, 1.1]} & 18.1 {\scriptsize[8.6, 27.6]} \\
        \bottomrule
    \end{tabular}
    \par\smallskip
    \parbox{\textwidth}{\footnotesize
    \textit{Note.}
    Recovery is evaluated on examples with \( |\Phi_r| \geq 0.5 \).
    Matched positions outside the demonstration blocks form the control.
    Drop metrics are computed as control minus targeted attenuation, so
    larger values indicate stronger dependence on the selected positions.}
    \vspace{-2pt}
\end{table}

Table~\ref{tab:demo_icl_main}(a) shows that all three models follow the induced mapping, while breaking the input--label pairing reduces mapping following with the same inputs and labels preserved in the prompt. This indicates that the rule is induced from relations established across the demonstrations. Table~\ref{tab:demo_icl_main}(b) shows that the inferred rule yields positive contribution margins and that substituting the demonstration positions recovers nearly the full effect, with negligible recovery from the matched control. Table~\ref{tab:demo_icl_main}(c) further shows that targeted attenuation reduces the label margin in all six settings and significantly lowers task accuracy in five, with the 95\% CI. Together, these results show that the inferred mapping produces a measurable role span contribution that affects the query decision. We next examine where this contribution is realized within the demonstrations.

\begin{table}[htbp]
    \centering
    \setlength{\abovecaptionskip}{2pt}
    \setlength{\belowcaptionskip}{3pt}
    \caption{\textbf{Position-wise organization of induced role effects.} Recovery and margin drop are compared across all demonstration tokens, label tokens, and input tokens.}
    \label{tab:demo_icl_localization}
    \small
    \setlength{\tabcolsep}{6pt}
    \begin{tabular}{@{}lccc@{\hspace{16pt}}ccc@{}}
        \toprule
        & \multicolumn{3}{c}{\textbf{(a) Recovery (\%)}} & \multicolumn{3}{c}{\textbf{(b) Margin drop vs.\ control }} \\
        \cmidrule(lr){2-4}\cmidrule(l){5-7}
        Model & All & Labels & Inputs & All & Labels & Inputs \\
        \midrule
        \multicolumn{7}{@{}l}{\emph{SST-2}} \\
        \hspace{0.8em}Qwen2.5-7B & 99.4 & 98.4 & 4.4 & 2.27 & 2.09 & 0.29 \\
        \hspace{0.8em}Mistral-7B & 99.8 & 98.7 & 2.4 & 1.24 & 1.37 & 0.38 \\
        \hspace{0.8em}Llama-3.1-8B & 96.2 & 95.1 & 3.2 & 0.63 & 0.76 & 0.23 \\
        \midrule
        \multicolumn{7}{@{}l}{\emph{TREC}} \\
        \hspace{0.8em}Qwen2.5-7B & 99.4 & 98.8 & 3.7 & 1.04 & 0.89 & 0.47 \\
        \hspace{0.8em}Mistral-7B & 98.5 & 91.1 & 11.1 & 1.37 & 1.35 & 0.14 \\
        \hspace{0.8em}Llama-3.1-8B & 97.0 & 90.9 & 1.6 & 0.69 & 0.49 & 0.10 \\
        \bottomrule
    \end{tabular}
    \par\smallskip
    \parbox{\textwidth}{\footnotesize
    \textit{Note.}
    Recovery is evaluated on examples with \( |\Phi_r| \geq 0.5 \).
    Matched positions outside the demonstration blocks form the control.
    Drop metrics are computed as control minus targeted attenuation, so
    larger values indicate stronger dependence on the selected positions.}
    \vspace{-2pt}
\end{table}

Table~\ref{tab:demo_icl_localization}(a) shows that the label positions recover nearly the full effect of the demonstration blocks. The input positions recover only a small fraction of this effect. Table~\ref{tab:demo_icl_localization}(b) shows the same organization under targeted attenuation. The margin drop at the label positions is close to that of all demonstration positions, while the input positions show a much smaller change. Together with the pairing result in Table~\ref{tab:demo_icl_main}(a), these findings separate rule formation from its direct realization at the query decision. The rule is induced from relations across demonstrations, and its direct contribution is concentrated at the label positions.

We further test whether failing label decisions can be restored by amplifying the demonstration contribution, using the same operation as in Section~\ref{sec:experiments-3-2}. Table~\ref{tab:icl_restoration}(a) shows that amplification restores a subset of these failing decisions, while random directions of equal norm do not at larger weights. Table~\ref{tab:icl_restoration}(b) further shows that, at similar deficits, single-label rule decisions recover less often than evidence decisions. Table~\ref{tab:icl_restoration}(c) shows that recovery follows the prediction in Appendix~\ref{app:decision-boundary}. Together, these results show that whether a failing decision can be restored depends on the support relative to the deficit.

\begin{table}[htbp]
    \centering
    \setlength{\abovecaptionskip}{2pt}
    \setlength{\belowcaptionskip}{3pt}
    \caption{\textbf{Restoration of failing single-label rule decisions in few-shot ICL.} \textbf{(a)} Target selection under amplification. \textbf{(b)} Recovery at matched deficits. \textbf{(c)} Recovery by the predicted threshold.}
    \label{tab:icl_restoration}
    \small
    \setlength{\tabcolsep}{3pt}
    \begin{tabular}{@{}lccc@{\hspace{9pt}}lc@{\hspace{9pt}}lc@{}}
        \toprule
        \multicolumn{4}{@{}c}{\textbf{(a) Selection (\%)}} & \multicolumn{2}{c}{\textbf{(b) Matched deficit (\%)}} & \multicolumn{2}{c@{}}{\textbf{(c) Prediction at $\beta{=}1$ (\%)}} \\
        \cmidrule(r){1-4}\cmidrule(lr){5-6}\cmidrule(l){7-8}
        Direction & $\beta{=}4$ & $\beta{=}8$ & $\beta{=}16$ & Group & Recovery & Group & Recovery \\
        \midrule
        Demonstration & \textbf{12.9} & \textbf{18.8} & \textbf{22.5} & Single-label rule & 6.5 & $\hat\beta_t\le 1$ & \textbf{59.5} \\
        Random & 5.5 & 0.0 & 0.0 & Task-defined evidence & 47.6 & $\hat\beta_t> 1$ & 0.4 \\
        & & & & Real-hallucinated & 58.8 & & \\
        \bottomrule
    \end{tabular}
    \par\smallskip
    \parbox{\textwidth}{\footnotesize
    \textit{Note.} \(\hat\beta_t\) follows Appendix~\ref{app:decision-boundary}.}
    \vspace{-2pt}
\end{table}

\subsection{Evidence from Multiple Sources}
\label{app:broader-evidence}
Our main evidence experiments place the relevant fact in a single support span, allowing us to isolate its effect on a factual choice. We next examine evidence use across multiple sources. One setting requires the model to combine information from several passages, while the other presents conflicting facts from two sources. In each setting, we identify which evidence positions affect the selected answer and test whether weakening them changes the factual choice.

\paragraph{Distributed evidence.} We first study questions from MuSiQue \citep{trivedi-etal-2022-musique} that require two reasoning steps. Each answer depends on a bridge passage and a passage that contains the final fact. The paired prompts preserve the question, passage order, and all other text, while changing only the answer entity. We use the union of the two passages as the evidence positions and matched distractor positions as the control. This construction lets us compare the evidence needed to establish the answer with the evidence that acts at the final choice.

\begin{table}[htbp]
    \centering
    \setlength{\abovecaptionskip}{2pt}
    \setlength{\belowcaptionskip}{3pt}
    \caption{\textbf{Evidence contributions under distributed multi-hop grounding.} \textbf{(a)} Grounded preference and pathway recovery. \textbf{(b)} Dependence on each supporting passage and its direct effect at the answer decision.}
    \label{tab:m3_distributed}
    \small
    \begin{tabular}{@{}l@{\hspace{8pt}}c@{\hspace{7pt}}c@{\hspace{7pt}}c@{\hspace{10pt}}c@{\hspace{10pt}}c@{}}
        \toprule
        \multicolumn{6}{c}{\textbf{(a) Grounded preference and pathway recovery}} \\
        \midrule
        & \multicolumn{3}{c}{Grounded preference (\%)} & \multicolumn{1}{c}{Matched effect} & \multicolumn{1}{c}{Pathway recovery} \\
        \cmidrule(lr){2-4}\cmidrule(lr){5-5}\cmidrule(l){6-6}
        Model & Original & Counterfactual & \shortstack{No evidence} & \shortstack{Median $M_e$} & \shortstack{Recovery\\difference (nats)} \\
        \midrule
        Qwen2.5-7B & 82.7 & 46.9 & 63.3 & 36.47 & 5.08 {\scriptsize[4.0, 6.2]} \\
        Mistral-7B & 89.8 & 43.9 & 69.4 & 35.11 & 5.07 {\scriptsize[4.2, 6.0]} \\
        Llama-3.1-8B & 98.0 & 50.0 & 76.5 & 26.08 & 4.30 {\scriptsize[3.6, 5.0]} \\
        \bottomrule
    \end{tabular}
    \par\vspace{5pt}
    \begin{tabular}{@{}l@{\hspace{7pt}}c@{\hspace{6pt}}c@{\hspace{9pt}}c@{\hspace{6pt}}c@{\hspace{6pt}}c@{\hspace{9pt}}c@{\hspace{6pt}}c@{}}
        \toprule
        \multicolumn{8}{c}{\textbf{(b) Support dependence and decision-level effects}} \\
        \midrule
        & \multicolumn{2}{c}{\shortstack{Prompt dependence:\\margin drop (nats)}} & \multicolumn{3}{c}{\shortstack{Decision effect:\\margin drop vs.\ control }} & \multicolumn{2}{c}{Token flip (\%)} \\
        \cmidrule(lr){2-3}\cmidrule(lr){4-6}\cmidrule(l){7-8}
        Model & \shortstack{Bridge\\removed} & \shortstack{Answer\\removed} & \shortstack{All evidence} & \shortstack{Answer span} & \shortstack{Bridge span} & Targeted & Control \\
        \midrule
        Qwen2.5-7B & \shortstack{1.98\\{\scriptsize[0.9, 3.1]}} & \shortstack{5.82\\{\scriptsize[4.5, 7.2]}} & \shortstack{5.89\\{\scriptsize[4.7, 7.1]}} & \shortstack{4.68\\{\scriptsize[3.7, 5.7]}} & \shortstack{0.09\\{\scriptsize[$-0.4$, 0.6]}} & 11.2 & 1.0 \\
        Mistral-7B & \shortstack{1.27\\{\scriptsize[0.7, 1.9]}} & \shortstack{5.46\\{\scriptsize[4.4, 6.6]}} & \shortstack{3.98\\{\scriptsize[3.2, 4.8]}} & \shortstack{3.48\\{\scriptsize[2.8, 4.2]}} & \shortstack{0.14\\{\scriptsize[$-0.1$, 0.4]}} & 8.2 & 0.0 \\
        Llama-3.1-8B & \shortstack{1.07\\{\scriptsize[0.6, 1.5]}} & \shortstack{4.50\\{\scriptsize[3.7, 5.3]}} & \shortstack{4.37\\{\scriptsize[3.6, 5.1]}} & \shortstack{3.42\\{\scriptsize[2.8, 4.0]}} & \shortstack{0.43\\{\scriptsize[0.1, 0.8]}} & 12.2 & 0.0 \\
        \bottomrule
    \end{tabular}
    \par\smallskip
    \begin{minipage}{0.95\columnwidth}
        \footnotesize\raggedright
        \textit{Note.} Recovery difference (nats) is the score recovery at the answer decision from evidence substitution relative to the matched distractor substitution.   Larger margin drop indicates stronger dependence.
    \end{minipage}
    \vspace{-2pt}
\end{table}

As shown in Table~\ref{tab:m3_distributed}(a), the matched effect favors the supported answer for all three models, and substitution through the evidence positions recovers most of this effect relative to the control. Removing either passage also reduces the answer margin, showing that both passages are needed to establish the answer (Table~\ref{tab:m3_distributed}(b)). Their direct effects at the final choice are not equal. Weakening the passage that contains the final fact causes a large margin drop in every model, whereas weakening the bridge passage has little direct effect in Qwen and Mistral and a much smaller effect in Llama. Thus, the answer depends on evidence distributed across the two passages, but the final choice depends mainly on the passage that supplies the final fact.

\paragraph{Competing evidence.} We next consider contexts in which two sources state different entities for the same fact. The prompt marks one source as primary. The paired prompts exchange the entities stated by the two sources while preserving the question, passage order, and remaining text. A closed-book condition measures the preference that remains without either claim. We then weaken the two sources separately to test how each one affects the same factual choice.

Table~\ref{tab:m3_competing}(a) shows that context-conditioned preference varies across settings. The original and closed-book rates are nearly identical for Qwen, separate more for Mistral, and differ most clearly for Llama. The matched effect remains positive in all three models. Table~\ref{tab:m3_competing}(b) reveals a more consistent source-specific effect. Weakening the primary source lowers the margin of its answer, whereas weakening the competing source can raise it. The primary intervention also changes the selected token far more often than the control. Thus, the models distinguish the effects of the two sources, even though the designated source does not exert the same control over the final answer.

Taken together, the two settings show that evidence dependence can be distributed across sources, while direct influence and final control remain source-specific. The context determines which information is required, that information shapes the factual choice, and the resulting influence may or may not determine the final answer. These findings extend the evidence contribution account beyond a single support span while preserving the distinction between contextual influence and final behavior.

\subsection{Role Span Influence around Key Decisions}
\label{app:role span-influence}

So far, our analysis shows that role span contributions act through the output decisions required by each role. The same pattern extends to rules inferred from demonstrations and evidence drawn from multiple sources in Appendices~\ref{app:demonstration-icl} and~\ref{app:broader-evidence}. For rules, these decisions govern structure and selection; for evidence, they determine factual content in the response. One may wonder how the role effect is organized around these semantic decisions. 

We answer this by comparing role span influence at the semantic decisions with matched output positions. As shown in Table~\ref{tab:d5_semantic_decisions}(a), role span influence is consistently higher at the semantic decisions than at the matched positions. The same concentration appears when the decisions are specified from reference responses. These results show that semantic decisions form distinct points of role span influence in the output.

Step further, we ask how the generation change is positioned around these semantic decisions. We measure the first divergence relative to each decision position. As shown in Table~\ref{tab:d5_semantic_decisions}(b), rule changes occur at or shortly before the decision itself, with \(44\%\)--\(53\%\) of first divergences occurring exactly at that position. In contrast, evidence shows a different pattern: \(84\%\)--\(87\%\) of first divergences occur before the factual value is emitted, with median offsets of several tokens. These results indicate that rule and evidence exhibit different positional patterns around their semantic decisions.

We further examine which pathway carries these earlier changes. On a fixed prefix, we remove only the direct evidence term at a position, or remove the remaining transport terms while keeping it. As shown in Table~\ref{tab:same_prefix_split}, the earlier first divergences occur at low-margin positions and are mainly carried by the indirect terms. In contrast, removing only the direct evidence term at the factual decision changes the selected fact far more often than the same removal on a matched non-evidence span. These results indicate that indirect transport shapes the tokens preceding the fact, while the factual decision itself remains sensitive to the direct evidence contribution.

In short, the semantic decisions provide distinct points of role span influence, while generation changes follow different positional patterns around them. Functional roles therefore organize both the output decisions that carry role span influence and the local pattern of generation changes around those decisions.

\begin{table}[htbp]
    \centering
    \setlength{\abovecaptionskip}{2pt}
    \setlength{\belowcaptionskip}{3pt}
    \caption{\textbf{Evidence contributions under competing factual sources.} \textbf{(a)} Context-conditioned preference and pathway recovery. \textbf{(b)} Source-specific effects under targeted attenuation.}
    \label{tab:m3_competing}
    \small
    \begin{tabular}{@{}l@{\hspace{8pt}}c@{\hspace{7pt}}c@{\hspace{7pt}}c@{\hspace{10pt}}c@{\hspace{7pt}}c@{\hspace{10pt}}c@{}}
        \toprule
        \multicolumn{7}{c}{\textbf{(a) Context-conditioned preference and pathway recovery}} \\
        \midrule
        & \multicolumn{3}{c}{Primary preference (\%)} & \multicolumn{2}{c}{Matched effect} & \multicolumn{1}{c}{Pathway recovery} \\
        \cmidrule(lr){2-4}\cmidrule(lr){5-6}\cmidrule(l){7-7}
        Model & Original & Swapped & \shortstack{Closed book} & \shortstack{Median $M_e$} & \shortstack{Positive $M_e$ (\%)} & \shortstack{Recovery difference (nats)} \\
        \midrule
        Qwen2.5-7B & 28.4 & 25.3 & 26.3 & 1.06 & 53.7 {\scriptsize[43.2, 64.2]} & 4.99 {\scriptsize[4.0, 6.0]} \\
        Mistral-7B & 31.6 & 13.7 & 22.1 & 1.90 & 65.3 {\scriptsize[55.8, 74.7]} & 4.63 {\scriptsize[3.9, 5.3]} \\
        Llama-3.1-8B & 34.7 & 5.3 & 20.0 & 4.03 & 81.1 {\scriptsize[72.6, 88.4]} & 4.70 {\scriptsize[4.1, 5.3]} \\
        \bottomrule
    \end{tabular}
    \par\vspace{5pt}
    \begin{tabular}{@{}l@{\hspace{10pt}}c@{\hspace{8pt}}c@{\hspace{11pt}}c@{\hspace{7pt}}c@{}}
        \toprule
        \multicolumn{5}{c}{\textbf{(b) Source-specific causal effects}} \\
        \midrule
        & \multicolumn{2}{c}{\shortstack{Source-specific decision effect }} & \multicolumn{2}{c}{Token flip (\%)} \\
        \cmidrule(lr){2-3}\cmidrule(l){4-5}
        Model & \shortstack{Primary drop} & \shortstack{Secondary gain} & Primary & Control \\
        \midrule
        Qwen2.5-7B & 6.43 {\scriptsize[5.2, 7.7]} & 5.30 {\scriptsize[4.2, 6.5]} & 23.2 & 3.2 \\
        Mistral-7B & 5.12 {\scriptsize[4.3, 5.9]} & 3.91 {\scriptsize[3.2, 4.6]} & 26.3 & 1.1 \\
        Llama-3.1-8B & 6.30 {\scriptsize[5.5, 7.1]} & 3.34 {\scriptsize[2.8, 3.9]} & 31.6 & 0.0 \\
        \bottomrule
    \end{tabular}
    \par\smallskip
    \begin{minipage}{0.95\columnwidth}
        \footnotesize\raggedright
        \textit{Note.} Primary drop is the margin decrease of the primary asserted candidate under targeted attenuation, relative to the matched control; Secondary gain is the margin increase after attenuating the competing source.
    \end{minipage}
    \vspace{-2pt}
\end{table}

\begin{table}[htbp]
    \centering
    \setlength{\abovecaptionskip}{2pt}
    \setlength{\belowcaptionskip}{3pt}
    \caption{\textbf{Role span behavior at semantic decisions.} \textbf{(a)} Role span influence at semantic decisions: the median within-response percentile of \(|\hat{\delta}_t|\) at the decision position and at matched positions in other responses. \textbf{(b)} Position of the first generation change relative to the semantic decision.}
    \label{tab:d5_semantic_decisions}
    \small
    \setlength{\tabcolsep}{4pt}
    \begin{tabular*}{\linewidth}{@{\extracolsep{\fill}}l c c c c c@{}}
        \toprule
        & \multicolumn{2}{c}{\textbf{(a) Role span influence (\%)}}
        & \multicolumn{3}{c}{\textbf{(b) Position of generation changes}} \\
        \cmidrule(lr){2-3}\cmidrule(l){4-6}
        Model
        & \shortstack{Decision percentile}
        & \shortstack{Matched control}
        & \shortstack{Median offset}
        & \shortstack{Before dec.\ (\%)}
        & \shortstack{At dec.\ (\%)} \\
        \midrule
        \multicolumn{6}{@{}l}{\emph{Rule}} \\
        \hspace{0.8em}Qwen2.5-7B & \textbf{85.0} & 67.4 & $+0$ & 46.9 & 53.1 \\
        \hspace{0.8em}Mistral-7B & \textbf{85.7} & 63.6 & $-1$ & 55.9 & 44.1 \\
        \hspace{0.8em}Llama-3.1-8B & \textbf{90.0} & 63.0 & $-5$ & 55.3 & 44.7 \\
        \midrule
        \multicolumn{6}{@{}l}{\emph{Evidence}} \\
        \hspace{0.8em}Qwen2.5-7B & \textbf{92.5} & 50.2 & $-8$ & 86.0 & 8.0 \\
        \hspace{0.8em}Mistral-7B & \textbf{77.5} & 51.2 & $-6$ & 87.2 & 4.3 \\
        \hspace{0.8em}Llama-3.1-8B & \textbf{95.0} & 50.5 & $-6$ & 83.7 & 9.3 \\
        \bottomrule
    \end{tabular*}
    \par\smallskip
    \parbox{\linewidth}{\footnotesize
    \textit{Note.} \textbf{(a)} Both columns show the median within-response percentile of \(|\hat{\delta}_t|\); Matched reads the same positions in other responses. \textbf{(b)} Offset is the first-divergence position minus the semantic-decision position (in tokens); negative values indicate an earlier change.}
    \vspace{-2pt}
\end{table}

\begin{table}[htbp]
    \centering
    \setlength{\abovecaptionskip}{2pt}
    \setlength{\belowcaptionskip}{3pt}
    \caption{\textbf{Direct and indirect evidence effects on a fixed prefix.} \textbf{(a)} Changes in the selected fact at the factual decision. \textbf{(b)} Pathway share and margin at the first divergence.}
    \label{tab:same_prefix_split}
    \small
    \setlength{\tabcolsep}{4pt}
    \begin{tabular*}{\linewidth}{@{\extracolsep{\fill}}l c c c c@{}}
        \toprule
        & \multicolumn{2}{c}{\textbf{(a) Factual decision (\%)}} & \multicolumn{2}{c}{\textbf{(b) First divergence}} \\
        \cmidrule(lr){2-3}\cmidrule(l){4-5}
        & \shortstack{Direct\\{\scriptsize evidence removed}} & \shortstack{Matched\\{\scriptsize non-evidence}} & \shortstack{Indirect share\\{\scriptsize median (\%)}} & \shortstack{Margin\\{\scriptsize median}} \\
        \midrule
        All models & \textbf{31.6} & 1.7 & 91.8 & 0.6 \\
        \bottomrule
    \end{tabular*}
    \par\smallskip
    \parbox{\linewidth}{\footnotesize
    \textit{Note.} The prefix is fixed to the original generation. Indirect terms follow the decomposition in Appendix~\ref{app:attention-contribution}.}
    \vspace{-2pt}
\end{table}

\subsection{Predicting Task Recovery}
\label{app:task_recovery_prediction}

Appendix~\ref{app:decision-boundary} predicts the recovery of each naturally failing decision under role contribution amplification. We first compare these predictions with the subsequent intervention outcomes. Across the two roles, the predicted and observed decision outcomes agree in 79\% of the cases, while the predicted amplification closely tracks the first observed crossing among recovered decisions.

Here, we show that the decision-level predictions can extend to task behavior. Following the role-specific decision organization identified in Section~\ref{sec:Role Span Influence}, we map the predicted decision outcomes to task recovery. For evidence, recovery of the factual choice determines the grounding prediction. For rules, the decision-level prediction is not directly comparable, since successful rule following requires sustained support across multiple structural and selection choices.

As shown in Table~\ref{tab:task_recovery_prediction}, the observed task recovery follows the predicted trend for evidence, while rule recovery stays near zero, as predicted. For evidence, task success increases from 25.2\% to 58.3\%, compared with a predicted increase from 23.3\% to 69.9\%. For rules, both remain near zero over the same range. Thus, we show that the first-order decision formulation can be used to quantitatively predict task recovery in both ICL and OOC hallucination.

\begin{table}[htbp] 
  \centering
  \setlength{\abovecaptionskip}{2pt}
  \setlength{\belowcaptionskip}{3pt}
  \setlength{\aboverulesep}{0.3ex}
  \setlength{\belowrulesep}{0.3ex}
  \caption{\textbf{Task-level recovery prediction under role contribution
  amplification.} Predicted versus observed task success across
  amplification strengths.}
  \label{tab:task_recovery_prediction}
  \small
  \begin{tabular}{@{}c@{\hspace{16pt}}c@{\hspace{12pt}}c@{\hspace{16pt}}c@{\hspace{12pt}}c@{}}
      \toprule
      & \multicolumn{2}{c}{Rule span}
      & \multicolumn{2}{c}{Evidence span} \\
      \cmidrule(lr){2-3}\cmidrule(l){4-5}
      $\beta$ & Predicted (\%) & Observed (\%)
      & Predicted (\%) & Observed (\%) \\
      \midrule
      0.5 & 0.0 & 0.0 & 23.3 & 25.2 \\
      1 & 0.0 & 1.3 & 26.2 & 32.0 \\
      2 & 0.0 & 2.5 & 38.8 & 47.6 \\
      4 & 0.0 & 6.3 & 52.4 & 53.4 \\
      8 & 0.0 & 3.8 & 64.1 & 55.3 \\
      16 & 6.3 & 6.3 & 69.9 & 58.3 \\
      \bottomrule
  \end{tabular}
  \par\smallskip
  \vspace{-2pt}
\end{table}

\end{document}